\documentclass{article}

\usepackage{times}
\usepackage{graphicx} 
\usepackage{subfigure}

\usepackage{natbib}

\usepackage{algorithm}
\usepackage{algorithmic}
\usepackage[english]{babel}
\usepackage{hyperref}

\usepackage[accepted]{icml2016arxiv}

\usepackage{amsmath,bm}
\usepackage{amsfonts}
\usepackage{color}
\usepackage{makecell}
\usepackage{multirow}
\usepackage{epstopdf}

\makeatletter
\renewcommand{\@thesubfigure}{}
\makeatother

\newcommand{\av}{\mathbf{a}} 
\newcommand{\bv}{\mathbf{b}}

 \newcommand{\Iv}{\mathbf{I}}

\newcommand{\pv}{\mathbf{p}} 
 
 \newcommand{\Rv}{\mathbf{R}}

\newcommand{\uv}{\mathbf{u}} \newcommand{\Uv}{\mathbf{U}}
 \newcommand{\Vv}{\mathbf{V}}
 \newcommand{\Wv}{\mathbf{W}}
\newcommand{\xv}{\mathbf{x}} \newcommand{\Xv}{\mathbf{X}}
 
\newcommand{\zv}{\mathbf{z}} \newcommand{\Zv}{\mathbf{Z}}

\newcommand{\alphav}{\boldsymbol{\alpha}} 
\newcommand{\betav}{\boldsymbol{\beta}}

\newcommand{\epsilonv}{\boldsymbol{\epsilon}}

\newcommand{\thetav}{\boldsymbol{\theta}}

\newcommand{\muv}{\boldsymbol{\mu}} 
 
\newcommand{\xiv}{\boldsymbol{\xi}}

\newcommand{\sigmav}{\boldsymbol{\sigma}}

\newcommand{\phiv}{\boldsymbol{\phi}}

\newcommand{\zerov}{\boldsymbol{0}}

\newcommand{\argmax}{\operatornamewithlimits{argmax}}
\newcommand{\ud}{\mathrm{d}}

\newcommand{\ep}{\mathbb{E}}

\newcommand{\N}{\mathcal{N}}

\newcommand{\RR}{\mathbb{R}}

\newcommand{\fig}[1]{Fig.~\ref{#1}}
\newcommand{\tabl}[1]{Table~\ref{#1}}
\newcommand{\eqn}[1]{Eqn.~(\ref{#1})}
\newcommand{\secref}[1]{Sec.~\ref{#1}}

\newcommand{\xvn}{\xv_{n}}
\newcommand{\zvn}{\zv_{n}}
\newcommand{\zvl}{\zv^{(l)}}
\newcommand{\vlb}{\mathcal{L}(\phiv;\thetav,\xv)}
\newcommand{\vll}{\mathcal{J}(\phiv;\thetav,\xv)}
\newcommand{\qz}{q(\zv|\xv;\phiv)}

\newcommand{\pmx}{p(\xv|\thetav)}
\newcommand{\pz}{p(\zv|\xv,\thetav)}

\newcommand{\recmodel}{recognition model}
\newcommand{\estll}{Est. LL.}
\newcommand{\methodg}{DSGNHT-Gibbs}
\newcommand{\methodq}{DSGNHT-NAIS}
\newcommand{\mnist}{MNIST}
\newcommand{\caltech}{Caltech 101 Silhouettes}
\newcommand{\omni}{Omniglot}
\newcommand{\splm}{Supplementary Material}
\newcommand{\darn}{DARN}

\icmltitlerunning{Learning Deep Generative Models with Doubly Stochastic MCMC}

\begin{document}

\twocolumn[
\icmltitle{Learning Deep Generative Models with Doubly Stochastic MCMC}

\icmlauthor{Chao Du}{du-c14@mails.tsinghua.edu.cn}
\icmlauthor{Jun Zhu}{dcszj@mail.tsinghua.edu.cn}
\icmlauthor{Bo Zhang}{dcszb@mail.tsinghua.edu.cn}
\icmladdress{Dept. of Comp. Sci. \& Tech., State Key Lab of Intell. Tech. \& Sys., TNList Lab, \\
Center for Bio-Inspired Computing Research, Tsinghua University, Beijing, 100084, China}

\icmlkeywords{Doubly Stochastic MCMC, machine learning, ICML}

\vskip 0.3in
]

\begin{abstract}
We present doubly stochastic gradient MCMC, a simple and generic method for (approximate) Bayesian inference of deep generative models (DGMs) in a collapsed continuous parameter space. At each MCMC sampling step, the algorithm randomly draws a mini-batch of data samples to estimate the gradient of log-posterior and further estimates the intractable expectation over hidden variables via a neural adaptive importance sampler, where the proposal distribution is parameterized by a deep neural network and learnt jointly. We demonstrate the effectiveness on learning various DGMs in a wide range of tasks, including density estimation, data generation and missing data imputation. Our method outperforms many state-of-the-art competitors.
\end{abstract}

\section{Introduction}

Learning deep models that consist of multi-layered representations has obtained state-of-the-art performance in many tasks~\cite{Bengio:14,hinton:06},
partly due to their ability on capturing high-level abstractions.
As an important family of deep models, deep generative models (DGMs)~\cite{hinton:06,Salakhutdinov:09} 
can answer a wide range of queries by performing probabilistic inference, such as inferring the missing values of input data, which is beyond the scope of recognition networks such as deep neural networks.

However, probabilistic inference with DGMs is challenging, especially when a Bayesian formalism is adopted,
which is desirable to protect the DGM from 
overfitting~\cite{MacKay:1992-Bayes,Neal:1995-Bayes} and to perform sparse Bayesian inference~\cite{Gan:aistats15} or nonparametric inference~\cite{Adams:2010} to learn the network structure. For Bayesian methods in general, the posterior inference often involves intractable integrals because of several potential factors, such as that the space is extremely high-dimensional and that the Bayesian model is non-conjugate. To address the challenges, approximate methods have to be adopted, including variational~\cite{jordan:1999,saul1996} and Markov chain Monte Carlo (MCMC) methods~\cite{Robert:2005}.

Much progress has been made on stochastic variational methods for DGMs~\cite{kingma14iclr,danilo14icml,Ranganath:2014}, under some mean-field or parameterization assumptions. One key feature of such variational methods is that they marry ideas from deep neural networks to parameterize the variational distribution by a recognition network and jointly learn the parameters by optimizing a variational bound.
In contrast, little work has been done on extending MCMC methods to learn DGMs in a Bayesian setting, which are often more accurate, except a few exceptions. \citet{Gan:aistats15} present a Gibbs sampler for deep sigmoid belief networks with a sparsity-inducing prior via data augmentation,
\citet{Adams:2010} present a Metropolis-Hastings method for cascading Indian buffet process and \citet{li2015high} develop a high-order stochastic
gradient MCMC method and apply to deep Poisson factor analysis~\cite{Gan:icml15}.


In this paper, we present a simple and generic method, named doubly stochastic gradient MCMC, to improve the efficiency of performing Bayesian inference on DGMs.
By drawing samples in the collapsed parameter space, our method extends the recent work on stochastic gradient MCMC~\cite{Welling:icml11,Ahn:2012,Chen:icml14,ding2014bayesian} to deal with the challenging task of posterior inference with DGMs. Besides the stochasticity of randomly
drawing a mini-batch of samples in stochastic approximation, our algorithm introduces an extra dimension of stochasticity to
estimate the intractable gradients by randomly drawing the hidden variables in DGMs. The sampling can be done via a Gibbs sampler, which however has a low mixing rate in high dimensional spaces. To address that, we develop a neural adaptive importance sampler (NAIS), where the adaptive proposal is parameterized by a recognition network and
the parameters are optimized by descending inclusive KL-divergence.
By combining the two types of stochasticity, we construct an asymptotically unbiased estimate of the gradient in the continuous parameter space.
Then, a stochastic gradient MCMC method is applied with guarantee to (approximately) converge to the target posterior when the learning rates are set under some proper annealing scheme.


Our method can be widely applied to the DGMs with either discrete or continuous hidden variables.
In experiments, we demonstrate the efficacy on learning various DGMs, such as deep sigmoid belief networks~\cite{Mnih:icml2014}, for density estimation, data generation and missing value imputation. Our results show that we can outperform many strong competitors for learning DGMs. 

\vspace{-.15cm}
\section{Related Work}
\vspace{-.15cm}


Recently, there has been a lot of interest in developing variational methods for DGMs.
One common strategy for dealing with the intractable posterior distribution is to approximate it with a recognition (or inference) network, and
a variational lower bound is then optimized~\cite{kingma14iclr,Mnih:icml2014}.
Note in these methods the gradients are also estimated doubly stochastically.
\citet{kingma14iclr} and \citet{Mnih:icml2014} adopt variance reduction techniques to make these methods practically applicable.
\citet{titsias14doubly} propose a so-called ``doubly stochastic variational inference'' method for non-conjugate Bayesian inference.
We are inspired by these methods when naming ours. 

The reweighted wake-sleep (RWS)~\cite{bornschein2014reweighted} and importance weighted autoencoder (IWAE)~\cite{burdaiwae} directly estimate the log-likelihood (as well as its gradient) via importance sampling, where the proposal distribution is characterized by a recognition model. These methods reduce the gap between the variational bound and the log-likelihood, which is shown much tighter than that in \citet{kingma14iclr}.
Such tighter bound results in an asymptotically unbiased estimator of its gradient.
We draw inspiration from these variational methods to build our MCMC samplers.

Our work is closely related to the recent progress on neural adaptive proposals for sequential Monte Carlo (NASMC)~\cite{GuGT15}.
Different from our work, NASMC deals with dynamical models
such as Hidden Markov models and adopts recurrent neural network as the proposal.
We use a similar KL-divergence as NASMC to learn the proposal. 

Finally, \citet{Gan:icml15} adopt a Monte Carlo estimate via Gibbs sampling to the intractable gradients under a stochastic MCMC method particularly for topic models.
Besides a general perspective which is applicable to various types of DGM models,
we propose a neural adaptive importance sampler which is more efficient than Gibbs sampling and leads to better estimates.


\vspace{-.15cm}
\section{Doubly Stochastic Gradient MCMC for Deep Generative Models}\label{sec:dsmcmc}
\vspace{-.15cm}

We now present the doubly stochastic gradient MCMC for deep generative models.

\vspace{-.15cm}
\subsection{Deep Generative Models}\label{sec:dsmcmc:dgm}
\vspace{-.15cm}

Let $\Xv = \{ \xvn \}^{N}_{n = 1}$ be a given dataset with $N$ i.i.d. samples.
A deep generative model (DGM) assumes that each sample $\xvn \in \RR^D$ is generated from a vector of hidden variables $\zvn \in \RR^H$,
which itself follows some prior distribution $p(\zv | \alphav)$.
Let $p(\xv | \zv, \betav)$ be the likelihood model.
The joint probability of a DGM is as follows:\\[-.3cm]
\begin{equation}\label{eq:DGM-joint-dist}
p(\Xv, \Zv| \thetav)  = \prod^{N}_{n = 1} p(\zvn | \alphav) p(\xvn | \zvn, \betav),
\end{equation}\\[-.3cm]
where $\thetav := (\alphav, \betav)$.
Depending on the structure of $\zv$, 
various DGMs have been developed, such as
deep belief networks~\cite{hinton:06},
deep sigmoid belief networks~\cite{Mnih:icml2014},
and deep Boltzmann machines~\cite{Salakhutdinov:09}.

For most DGMs,
the hidden variables $\zv$ are often assumed to have a directed multi-layer representation $\zv=\{\zvl\}_{l=1}^L$,
where $L$ is the number of hidden layers. Then the prior distribution has the factorization form:\\[-.4cm]
\begin{equation}\label{eq:prior-deep}
p(\zv | \thetav)  = p(\zv^{(L)} | \thetav) \prod^{L-1}_{l = 1} p(\zvl | \zv^{(l+1)}, \thetav),
\end{equation}\\[-.4cm]
where $p(\zvl | \zv^{(l+1)}, \thetav)$ is defined by some conditional stochastic layer that
takes $\zv^{(l+1)}$ as input and generates samples $\zvl$.
The likelihood model is further assumed to be a conditional stochastic layer again:\\[-.3cm]
\begin{equation}\label{eq:likelihood-deep}
p(\xv | \zv, \thetav)  = p(\xv |\zv^{(1)}, \thetav),
\end{equation}\\[-.5cm]
where the samples are generated conditioned on the lowest hidden layer only.

Various conditional stochastic layers have been developed. 
In the following we briefly summarize the layers used in our experiments:

\textbf{Sigmoid Belief Network layer (SBN):}
A SBN layer~\cite{saul1996} is a directed graphical model that defines the conditional probability of each independent binary variable $z^{(l)}_i$
given the upper layer $\zv^{(l+1)}$ as follows:\\[-.3cm]
\begin{equation}\label{eq:prob-sbn}
p(z^{(l)}_i = 1|\zv^{(l+1)}) = \sigma(\Wv_{i,:}\zv^{(l+1)}+b_i),
\end{equation}\\[-.5cm]
where $\sigma(x)\!=\!1/(1+e^{-x})$ is the sigmoid function, $\Wv_{i,:}$ denotes the $i$-th row of the weight matrix and $b_i$ is the bias.

\textbf{Deep Autoregressive Network layer (\darn{}):} A \darn{}~\cite{gregor14deep} layer assumes in-layer connections on the SBN layer.
It defines the probability of each binary variable $z^{(l)}_i$ conditioned on both the upper layer $\zv^{(l+1)}$ and the previous $\zv^{(l)}_{<i}$ in the same layer:\\[-.2cm]
\begin{equation*}\label{eq:prob-darn}
\vspace{-.1cm}
\begin{aligned}
p(z^{(l)}_i = 1|\zv^{(l)}_{<i}, \zv^{(l+1)}) = \sigma(\Uv_{i,:}\zv^{(l+1)}+\Wv_{i,<i}\zv^{(l)}_{<i}+b_i),
\end{aligned}
\end{equation*}\\[-.2cm]
where $\zv^{(l)}_{<i}$ refers to $(z^{(l)}_{1},\cdots,z^{(l)}_{i-1})^\top$ and $\Wv_{i,<i}$ denotes the first $i$ elements of the $i$-th row of the in-layer connection weight matrix.

\textbf{Conditional NADE layer:} The NADE~\cite{Larochelle:11}
models the distribution of high-dimensional discrete variables $\xv$ autoregressively with an internal MLP~\cite{bengio1999modeling}.
The dependency between the variables is captured by a single-hidden-layer feed-forward neural network:\\[-.2cm]
\begin{equation*}
p(x_i=1|\xv_{<i})=\sigma(\Vv_{i,:}\sigma(\Wv_{:,<i}\xv_{<i}+\av)+b_i).
\end{equation*}\\[-.4cm]
\citet{boulanger2012modeling} and \citet{bornschein2014reweighted} amend this model to a conditional NADE layer:\\[-.4cm]
\begin{equation}\label{eq:prob-nade}
\begin{aligned}
p(z^{(l)}_i = 1|& \zv^{(l)}_{<i}, \zv^{(l+1)}) = \sigma\Big(\Vv_{i,:}\sigma(\Wv_{:,<i}\zv^{(l)}_{<i}+  \\
&\Uv\zv^{(l+1)}+\av)+\Rv_{i,:}\zv^{(l+1)}+b_i\Big),
\end{aligned}
\end{equation}\\[-.3cm]
where we use $\Wv_{:,<i}$ to refer the sub-matrix consisting the first $i$ columns of $\Wv$.

\textbf{Variational Auto-Encoder layer (VAE):} VAE~\cite{kingma14iclr} differs from the above layers in that its output can be binary or real-valued variables.
It contains an internal MLP $f(\zv^{(l+1)})$ which encodes the parameters of the distribution $p(\zvl|\zv^{(l+1)})$.
The MLP may itself contain multiple deterministic layers. For binary output variable, the distribution of each individual variable is:\\[-.2cm]
\begin{equation}\label{eq:prob-vae-bin}
p(z^{(l)}_i=1|\zv^{(l+1)}) = \sigma(\Wv_{i,:}f(\zv^{(l+1)}) + b_i).
\end{equation}\\[-.4cm]
For real-value output variable,
the distribution of each independent variable $z^{(l)}_i$ is a normal distribution whose mean and variance are as follows:\\[-.2cm]
\begin{equation}
\vspace{-.1cm}
\begin{aligned}\label{eq:prob-vae-real}
\mu_i =& \Wv_{{\scriptscriptstyle\muv}i,:}f(\zv^{(l+1)}) + b_{{\scriptscriptstyle\muv} i}, \\
\log \sigma_i^2 =& \Wv_{{\scriptscriptstyle\sigmav}i,:}f(\zv^{(l+1)}) + b_{{\scriptscriptstyle\sigmav} i}.
\end{aligned}
\end{equation} 

\textbf{Top layer and Likelihood model:} The likelihood model in \eqn{eq:likelihood-deep} can be obtained by treating $\xv$ as $\zv^{(0)}$.
The distribution of top layer $p(\zv^{(L)} | \thetav)$, which has no ancestral layer,
can be obtained by simply treating the input as $\zerov$ vector
or setting to fixed distribution, e.g., standard normal for the VAE layer~\cite{kingma14iclr}. Detailed description of the layers and the model construction can be found in \splm{}.

\vspace{-.15cm}
\subsection{Variational MLE for DGMs}\label{sec:dsmcmc:variational}
\vspace{-.15cm}

Learning DGMs is often very challenging due to the intractability of posterior inference.
One popular type of methods resort to stochastic variational methods under the maximum likelihood estimation (MLE) framework,
$\hat \thetav = \argmax_{\thetav} \log p(\Xv | \thetav)$.
These methods commonly utilize some variational distribution $q(\zv|\xv;\phiv)$ to approximate the true posterior $p(\zv | \xv, \thetav)$.
For DGMs, the variational distribution $q(\zv|\xv;\phiv)$ can be formalized as a \recmodel{} (or inference network)
\cite{kingma14iclr,Mnih:icml2014,bornschein2014reweighted},
which takes $\xv$ as inputs and outputs $\zv$ stochastically.
Specifically, for the DGMs with multi-layer representation $\zv=\{\zvl\}_{l=1}^L$ described in \secref{sec:dsmcmc:dgm},
the variational distribution can be formulated as:\\[-.4cm]
\begin{equation}\label{eq:prob-variationdist}
q(\zv |\xv, \phiv)  = q(\zv^{(1)} |\xv, \phiv) \prod^{L-1}_{l = 1} q(\zv^{(l+1)} |\zvl , \phiv),
\end{equation}\\[-.4cm]
where each $q(\zv^{(l+1)} |\zvl , \phiv)$ and $q(\zv^{(1)} |\xv, \phiv)$ are again defined by some stochastic layers parametrized by $\phiv$.
With the variational distribution, a variational bound of the log-likelihood $\log p(\Xv | \thetav)$ can be derived and optimized,
e.g., the variational lower bound in \cite{kingma14iclr} and a tighter bound in \cite{burdaiwae}.

However, the variational bound
is often intractable to compute analytically for DGMs. To address this challenge,
recent progress~\cite{kingma14iclr,danilo14icml,Mnih:icml2014} has adopted hybrid Monte Carlo and variational methods, which
approximate the intractable expectations and their gradients over the parameters $(\thetav, \phiv)$ via
some unbiased Monte Carlo estimates. Furthermore, to handle large-scale datasets,
stochastic optimization~\cite{Robbins:1951,Bottou:1998} of the variational objective can be used with a suitable learning
rate annealing scheme. Note variance reduction is a key part of these methods to have fast and stable convergence.

\vspace{-.15cm}
\subsection{Doubly Stochastic Gradient MCMC}\label{sec:dsmcmc:dsmcmc}
\vspace{-.15cm}


We consider the Bayesian setting to 
infer the posterior distribution $p(\thetav, \Zv | \Xv) \propto p_0(\thetav) p(\Zv | \thetav) p(\Xv |\Zv, \thetav)$ or its marginal distribution $p(\thetav | \Xv)$,
by assuming some prior $p_0(\thetav)$. A Bayesian formalism of deep learning enjoys several advantages, such as preventing the model from overfitting and performing sparse/nonparametric Bayesian inference, as mentioned before.
However, except a handful of special examples, the posterior distribution is intractable to infer. 
Though variational methods can be developed as in~\cite{kingma14iclr,danilo14icml,Mnih:icml2014,bornschein2014reweighted}, under some mean-field or parameterization assumptions, 
they often require non-trivial model-specific deviations and may lead to inaccurate approximation when the assumptions are not properly made. Here, we consider MCMC methods, which are more generally applicable and can asymptotically approach the target posterior.

A straightforward application of MCMC methods can be Gibbs sampling or stochastic gradient MCMC~\cite{Welling:icml11,Ahn:2012,Chen:icml14,ding2014bayesian}. However, a Gibbs sampler can suffer from the random-walk behavior in high-dimensional spaces. Furthermore, a Gibbs sampler would need to process all data at each iteration, which is prohibitive when dealing with large-scale datasets. The stochastic gradient MCMC methods can lead to significant speedup by exploring statistical redundancy in large datasets; but they require that the sample space is continuous, which is not true for many DGMs, such as deep sigmoid belief networks that have discrete hidden variables. Below, we present a doubly stochastic gradient MCMC with general applicability.

\vspace{-.15cm}
\subsubsection{General Procedure}
\vspace{-.15cm}

We make the mildest assumption that the parameter space is continuous and the log joint distribution
$\log p(\xv, \zv | \thetav)$ is differentiable with respect to the model parameters $\thetav$ almost everywhere except a zero-mass set.
Such an assumption is true for almost all existing DGMs.
Then, our method draws samples in a collapsed space that involves the model parameters $\thetav$ only, by integrating out the hidden variables $\zv$:\\[-.6cm]
\begin{eqnarray}
p(\thetav | \Xv) = \frac{1}{p(\Xv)} p_0(\thetav) \prod_{n=1}^N \int p(\xv_n, \zv_n | \thetav)~\ud \zv_n,
\end{eqnarray}\\[-.4cm]
where for discrete variables the integral will be a summation.
Then the gradient of the log-posterior is 
$
\nabla_{\thetav} \log p(\thetav | \Xv) = \nabla_{\thetav} \log p_0(\thetav) + \sum_{n=1}^N \nabla_{\thetav} \log p(\xv_n |\thetav), 
$ 
where the second term can be calculated as:\\[-.5cm] 
\begin{align}\label{eq:derivative}
\nabla_{\thetav} \log \pmx
&= \frac{1}{\pmx}\frac{\partial}{\partial \thetav}\int p(\xv, \zv | \thetav)~\ud\zv \nonumber \\
&= \int \frac{p(\xv, \zv | \thetav)}{\pmx} \frac{\partial}{\partial \thetav} \log p(\xv, \zv | \thetav)~\ud\zv \nonumber \\
&= \ep_{\pz}\left[\frac{\partial}{\partial \thetav} \log p(\xv, \zv | \thetav)\right].
\end{align}\\[-.4cm]
With the above gradient, we can adopt a stochastic gradient MCMC (SG-MCMC) method to draw samples of $\thetav$.
We consider the stochastic gradient Nos\'{e}-Hoover thermostat (SGNHT)~\cite{ding2014bayesian}.
Note our method can be naturally extended to other SG-MCMC methods, e.g.,
stochastic gradient Langevin dynamics~\cite{Welling:icml11},
stochastic gradient Hamiltonian Monte Carlo~\cite{Chen:icml14} and
high-order stochastic gradient thermostats~\cite{li2015high}.
SGNHT defines a potential energy $U(\thetav)=-\log p(\thetav | \Xv)$ where $p(\thetav | \Xv)$ is the target posterior distribution,
and use a random mini-batch $B$ of the data $\Xv$ to approximate the true gradient of the potential energy: 
\begin{equation}
\!\nabla_{\thetav} \tilde{U}(\thetav)\!=\! - \nabla_{\thetav} \log p_0(\thetav) \!-\! \frac{N}{|B|} \! \sum_{n\in B}\! \nabla_{\thetav} \log p(\xv_n |\thetav).
\end{equation}\\[-.3cm]
We follow \citet{Gan:icml15} to use the multivariate version of SGNHT that generate samples by simulating the dynamics as follows:\\[-.5cm]
\begin{align}
\thetav_{t+1} &= \thetav_{t} + \lambda\pv_{t}, \label{eq:sgnht} \\
\pv_{t+1} &= \pv_{t} - \lambda\xiv_{t}\odot\pv_{t}-\lambda\nabla_{\thetav} \tilde{U}(\thetav_{t+1}) + \sqrt{2A}\N(\zerov,\lambda\Iv), \nonumber \\
\xiv_{t+1} &= \xiv_{t} + \lambda(\pv_{t+1}\odot\pv_{t+1}-\Iv), \nonumber
\end{align}\\[-.5cm]
where $\odot$ represent element-wise product, $\pv$ are the augmented momentum variables, $\xiv$ are the diffusion factors, $\lambda$ is the step size and $A$ is a constant that
controls the noise injected.
With a proper annealing scheme over the step size $\lambda$, the Hamiltonian dynamics will converge to the target posterior.

\vspace{-.15cm}
\subsubsection{Neural Adaptive Importance Sampler}
\vspace{-.15cm}

The remaining challenge is to compute the gradient as the expectation in \eqn{eq:derivative} is often intractable for DGMs.
Here, we construct an unbiased estimate of the gradient by a set of samples
$\{\zv^{(s)}\}_{s=1}^S$ from the posterior $p(\zv | \xv, \thetav)$:\\[-.5cm]
\begin{eqnarray}\label{eq:gradient-gibbs}
\nabla_{\thetav} \log \pmx 
\approx \frac{1}{S} \sum_{s=1}^S \left( \frac{\partial}{\partial \thetav} \log p(\xv, \zv^{(s)} | \thetav) \right).
\end{eqnarray}\\[-.4cm]
To draw the samples $\zv^{(s)}$,
a straightforward strategy is Gibbs sampling. Gibbs samplers are simple and applicable to both discrete and continuous hidden variables.
However, it may be hard to develop Gibbs samplers for most DGMs, as the highly complicated models often result in non-conjugacy.
More importantly, a Gibbs sampler can be slow to mix in high-dimensional spaces.
Below, we present a neural adaptive importance sampler (NAIS),
which again applies to both discrete and continuous hidden variables but with faster mixing rates.

Let $\qz$ be a proposal distribution which satisfies $\qz>0$ wherever $p(\zv | \xv, \thetav)>0$, we then have
\begin{align}
\nabla_{\thetav} \log \pmx = \ep_{\qz}\left[\frac{\pz}{\qz}\frac{\partial}{\partial \thetav} \log p(\xv, \zv | \thetav)\right], \nonumber
\end{align}
from which an unbiased importance sampling estimator can be derived with the sample weights being $\frac{\pz}{\qz}$.
However, computing $\pz$ is often hard for most DGMs. 
By noticing that $\pz \propto p(\xv, \zv | \thetav)$ and computing $p(\xv, \zv | \thetav)$ is easy, we derive a self-normalized importance sampling estimate as follows:
\begin{eqnarray}\label{eq:gradient-selfnormalize}
\nabla_{\thetav} \log \pmx
\approx  \frac{\sum_{s=1}^{S}\left( \frac{\partial}{\partial \thetav} \log p(\xv, \zv^{(s)} | \thetav)\right)\cdot\omega^{(s)}}{\sum_{s=1}^{S}\omega^{(s)}},
\end{eqnarray}
where $\{\zv^{(s)}\}_{s=1}^S$ is a set of samples drawn from the proposal $\qz$
and $\omega^{(s)}=\frac{p(\xv,\zv^{(s)}| \thetav)}{q(\zv^{(s)}|\xv;\phiv)}$ is the unnormalized likelihood ratio.
This estimate is asymptotically consistent \cite{mcbook}, and its slight bias decreases as drawing more samples.

\textbf{Neural Adaptive Proposals}: To reduce the variance of the estimator in \eqn{eq:gradient-selfnormalize} and get accurate gradient estimates,
$\qz$ should be as close to $\pz$ as possible.
Here, we draw inspirations from variational methods and learn adaptive proposals \cite{GuGT15} by minimizing some criterion. Specifically,
we build a \recmodel{} (or inference network) to represent the proposal distribution $\qz$ of hidden variables,
as in the variational methods \cite{kingma14iclr,bornschein2014reweighted}.
Such a \recmodel{} takes $\xv$ as input and outputs $\{\zv^{(s)}\}$ as samples from $\qz$, as described in \secref{sec:dsmcmc:variational}.
We optimize the quality of the proposal distribution by minimizing the
inclusive KL-divergence between the target posterior distribution and the proposal
$\ep_{\pz}[\log\frac{\pz}{\qz}]$ \cite{bornschein2014reweighted,GuGT15} or equivalently maximizing
the expected log-likelihood of the \recmodel{}
\begin{equation}\label{eq:eplikelihood}
\vll = \ep_{\pz}[\log\qz].
\end{equation}
We choose this objective due to the following reasons. If the target posterior belongs to the family of proposal distributions, maximizing $\vll$ leads to the optimal solution that is the target posterior;
otherwise, minimizing the inclusive KL-divergence tends to find proposal distributions that have higher entropy than the target posterior.
Such a property is advantageous for importance sampling as we require that $\qz > 0$ wherever $\pz > 0$. In contrast, the exclusive KL-divergence $\vlb := \ep_{\qz}[\log\frac{\qz}{\pz}]$, as widely adopted in the variational methods~\cite{kingma14iclr,danilo14icml,Mnih:icml2014}, does not have such a property --- It can happen that $\qz = 0$ when $\pz > 0$; therefore unsuitable for importance sampling.

The gradient of $\vll$ with respect to the parameters of the proposal distribution is
\begin{eqnarray}
\nabla_{\phiv} \vll = \ep_{\pz}[\nabla_{\phiv} \log \qz],
\end{eqnarray}
which can be estimated using importance sampling similar as in \eqn{eq:gradient-selfnormalize}:\\[-.5cm]
\begin{eqnarray}\label{eq:gradient-recmodel}
\nabla_{\phiv} \vll
\approx  \frac{\sum_{s=1}^{S}\left( \frac{\partial}{\partial \phiv} \log q(\zv^{(s)} ;\xv, \thetav)\right)\cdot\omega^{(s)}}{\sum_{s=1}^{S}\omega^{(s)}},
\end{eqnarray}\\[-.4cm]
where $\{\zv^{(s)}\}_{s=1}^{S}$ are samples from the latest proposal distribution $\qz$ and the weights are the same as in \eqn{eq:gradient-selfnormalize}.
To improve the efficiency, we adopt stochastic gradient descent methods to optimize the objective $\mathcal{J}(\phiv;\thetav,\Xv) := \sum_{n=1}^N \mathcal{J}(\phiv;\thetav,\xvn)$, with the gradient being estimated by a random mini-batch of data points $B$ at each iteration:\\[-.5cm]
\begin{eqnarray}\label{eq:gradient-recmodel-batch}
\nabla_{\phiv} \mathcal{J}(\phiv;\thetav,\Xv) \approx\frac{N}{|B|} \sum_{n \in B} \nabla_{\phiv} \mathcal{J}(\phiv;\thetav,\xvn),
\end{eqnarray}\\[-.35cm]
where each term $ \nabla_{\phiv} \mathcal{J}(\phiv;\thetav,\xvn)$ is further estimated by samples as in \eqn{eq:gradient-recmodel}.

With the above gradient estimates, we get the overall algorithm with neural adaptive importance sampling,
as outlined in Alg.~\ref{algo:dsmcmc},
where we adaptively update the proposal distribution by performing one step of \recmodel{} update after each step of SGNGT simulation.
Practically, re-sampling the hidden variables before each updating is helpful to get more accurate estimations.
The more detailed version of Alg.~\ref{algo:dsmcmc} is included in \splm{}.

\begin{algorithm}[tb]
   \caption{Doubly Stochastic Gradient MCMC with Neural Adaptive Proposals}
   \label{algo:dsmcmc}
\begin{algorithmic}
   \STATE {\bfseries Input:} data $\Xv$
   \STATE Initialize $\thetav$, $\phiv$
   \FOR{epoch $=1,2,\cdots$}
   \FOR{mini-batch $B_i \subset \{1, \cdots, N\}$}
   \STATE Sample $\{\zv_n^{(s)}\} \sim q(\zv|\xvn; \phiv)$, $n \in B_i$
   \STATE Estimate $\nabla \log p(\xvn | \thetav)$ with \eqn{eq:gradient-selfnormalize}, $n \in B_i$
   \STATE Update $\thetav$ with \eqn{eq:sgnht}
   \STATE Sample $\{\zv_n^{(s)}\} \sim q(\zv|\xvn; \phiv)$, $n \in B_i$ (optionally)
   \STATE Update $\phiv$ with the gradient in \eqn{eq:gradient-recmodel-batch}
   \ENDFOR
   \ENDFOR
   \STATE {\bfseries Output:} samples of $\thetav$
\end{algorithmic}
\end{algorithm}

\vspace{-.1cm}
\section{Experiments}\label{sec:exp}
\vspace{-.1cm}

We now present a series of experimental results of our doubly stochastic MCMC method on several representative deep generative models.
We use the doubly stochastic gradient Nos\'{e}-Hoover thermostat with a neural adaptive importance sampler (\methodq{}) in the experiments.
In \secref{sec:exp:discrete}, various DGMs with discrete hidden variables, such as sigmoid belief networks~\cite{DBLP:journals/ai/Neal92},
are trained on
the binarized {\bf \mnist{}}~\cite{Salakhutdinov08on} and the {\bf \caltech{}}~\cite{marlin2010inductive} datasets.
We compare the predictive performance with state-of-the-art methods in terms of the estimated log-likelihood (Est. LL.) on the test set. We also demonstrate the generative performance and analyze the sensitivity to main hyperparameters.
In \secref{sec:exp:continuous}, we train variational auto-encoders~\cite{kingma14iclr} on the binarized \mnist{} and the {\bf \omni{}}~\cite{lake2013one} datasets.

\begin{table*}[tb]\vspace{-.3cm}
\caption{MNIST results of various methods on five benchmark architectures.
``Dim'' denotes the number of hidden variables in each layer, with layer closest to the data laying left.
Values within brackets are variational lower bounds, values without brackets are estimated log-likelihoods.
$(\star)$ Use NADE layers for \recmodel{}.
The results of NVIL are from \citet{Mnih:icml2014};
the results of Wake-sleep and RWS are from \citet{bornschein2014reweighted};
and the results of Data Augmentation (DA) are from \citet{Gan:aistats15}.}
\label{table:logpvs-methods}
\vspace{-0.08cm}
\begin{center}
\begin{small}
\begin{tabular}{llcccccc}
\hline
Model     & Dim         &   NVIL      &  Wake-Sleep      &     RWS         &  DA         & \methodg{} & \methodq{}             \\
\hline
SBN       & $200$       & $(-113.1) $ & $-116.3(-120.7)$ &    $-103.1 $    & $(-113.02)$ & $-102.9$   & $\mathbf{-101.8}$      \\
SBN       & $200$-$200$ & $(-99.8)  $ & $-106.9(-109.4)$ &    $-93.4  $    & $(-110.74)$ & $-100.6$   & $\mathbf{-92.5}$       \\
SBN  & $200$-$200$-$200$& $(-96.7)  $ & $-101.3(-104.4)$ &    $-90.1  $    &     --      & $-97.5 $   & $\mathbf{-89.9}$       \\
\darn{}   & $200$       &     --      &        --   &$\mathbf{-89.2}^\star$& $(-102.11)$ & $-101.1$   & $-89.3^\star$          \\
NADE      & $200$       &     --      &        --        &    $-86.8^\star$&     --      &     --     & $\mathbf{-83.7}^\star$ \\
\hline
\end{tabular}
\end{small}
\end{center}
\vspace{-.4cm}
\end{table*}

\begin{figure}[tb]\vspace{-.3cm}
\begin{center}
\includegraphics[width=0.50\textwidth]{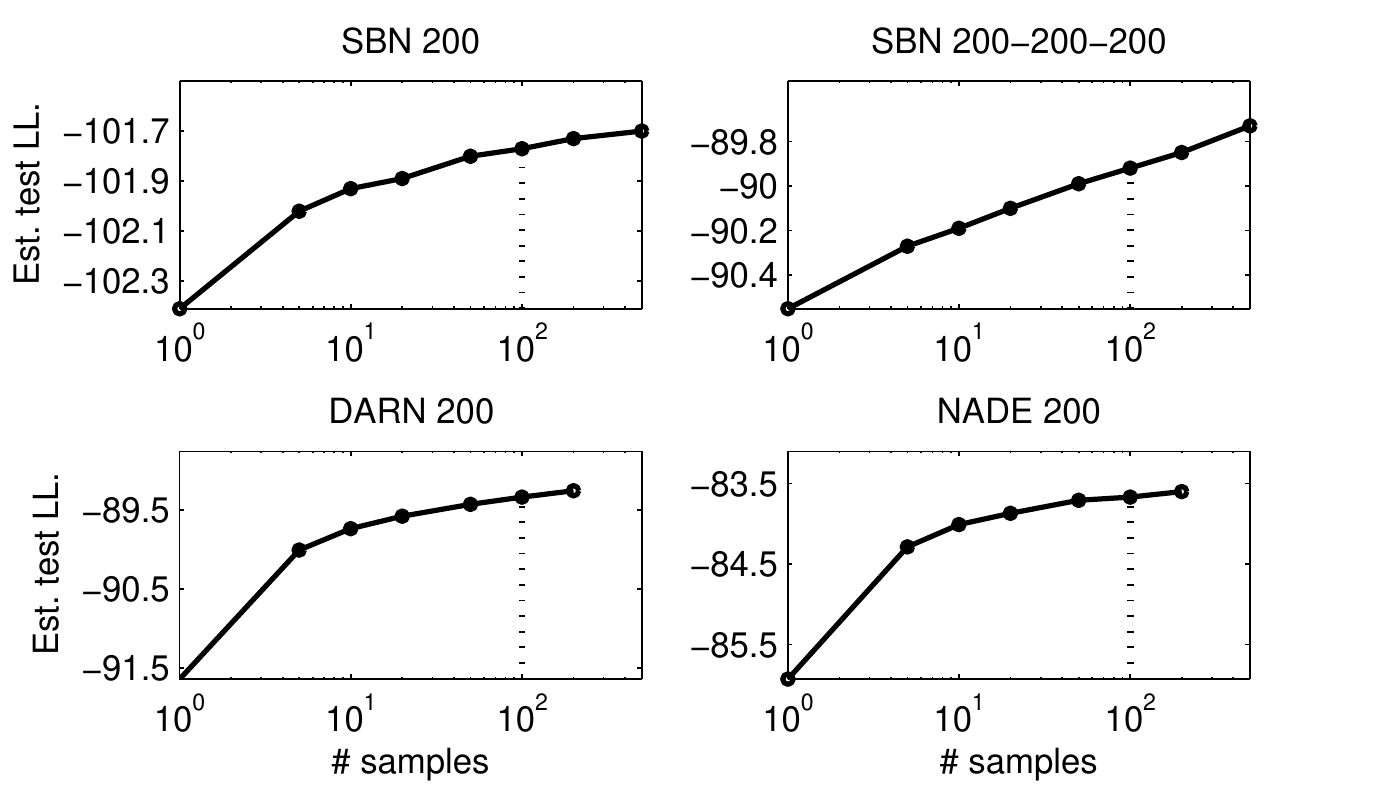}
\end{center}\vspace{-.5cm}
\caption{Log-likelihood estimation on \mnist{} for different models w.r.t number of posterior samples $M$ used for posterior mean estimator. Dotted line marks the results reported in \tabl{table:logpvs-methods}.}
\label{fig:post-samples}
\vskip -.4cm
\end{figure}

\vspace{-.1cm}
\subsection{Setup}
\vspace{-.1cm}

In our experiments, all models (including \recmodel{}s) are initialized following the heuristic of~\citet{glorot2010understanding}.
We set the Student-t prior to all the model parameters.
We use the reformulated form of multivariate SGNHT as described in \splm{}.
The per-batch learning rate $\gamma$ is set among \{0.01, 0.005, 0.001\}, 
from which we report the experiment with best performance. 
If not noted otherwise, the number of samples used during training is set to $S = 5$.
The mini-batch size $|B|$ is set to $100$ for all experiments.
The parameters of \recmodel{} are updated using
the Adam \cite{kingma15adam} optimizer with step sizes of $\{1,3,5\}\times 10^{-4}$. 

As our method infers the posterior $p(\thetav|\Xv)$, we adopt the posterior mean estimator for model evaluation:\vspace{-.3cm}
\begin{equation}\label{eq:posterior-mean}
\hat{\thetav}=\ep_{p(\thetav|\Xv)}[\thetav]\approx\frac{1}{M}\sum_{m=1}^{M}\thetav^{(m)}.\vspace{-.1cm}
\end{equation}
To compute the posterior mean,
we start to collect posterior samples when we observe the Est. LL. on the validation set does not increase for 10 consecutive epochs.
Then $M$ samples $\{\thetav^{(m)}\}_{m=1}^M$ from $M$ more epochs are averaged for final evaluation.
If not mentioned otherwise, the number of samples used for computing the posterior mean is set to $M = 100$.
We will also show how $M$ influences the results.

To evaluate the inferred model $\hat{\thetav}$ in terms of \estll{}, we adopt the $K$-sample importance weighting estimation $\mathcal{L}_K$:\\[-.5cm]
\begin{equation}\label{eq:estimation}
\mathcal{L}_K = \ep_{\zv^{(k)}\sim\qz}\left[\log\frac{1}{K}\sum_{k=1}^{K}\frac{p(\xv,\zv^{(k)})}{q(\zv^{(k)}|\xv)}\right].\vspace{-.1cm}
\end{equation}
Such estimation is also used by \citet{bornschein2014reweighted} and \citet{burdaiwae}.
We will clarify how we set the number of samples $K$ used for estimating the log-likelihoods and
investigate how it influences the quality of the estimator.

\darn{} and NADE are not permutation-invariant models. In our experiments, the ordering is simply determined by the original order in the dataset.

See \splm{} for more details about the experimental setting.

\begin{figure}[tb]\vspace{-.3cm}
\begin{center}
\subfigure{
\label{fig:train-samples}
\includegraphics[width=0.22\textwidth]{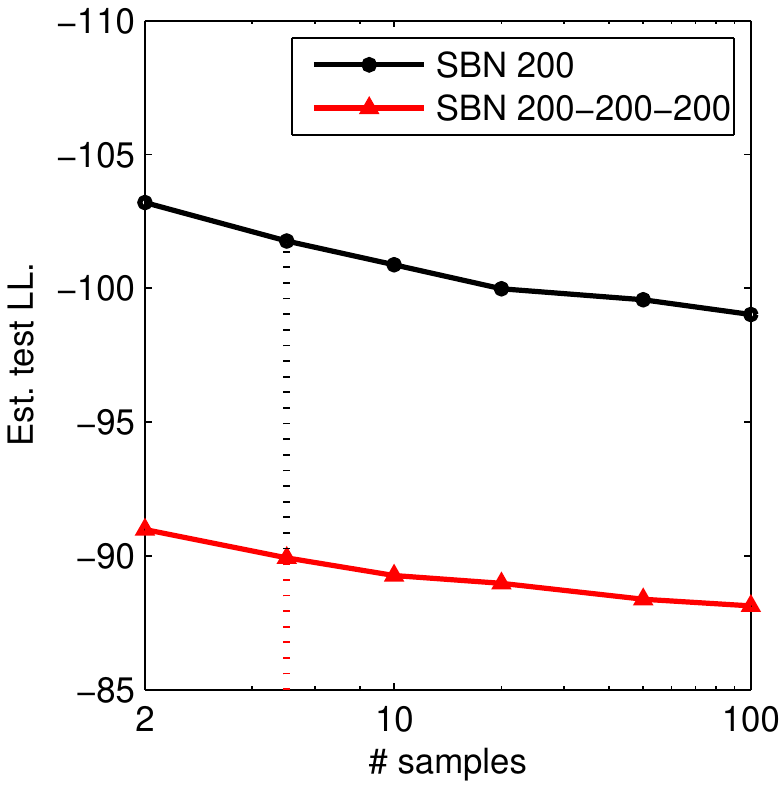}
}
\subfigure{
\label{fig:test-samples}
\includegraphics[width=0.22\textwidth]{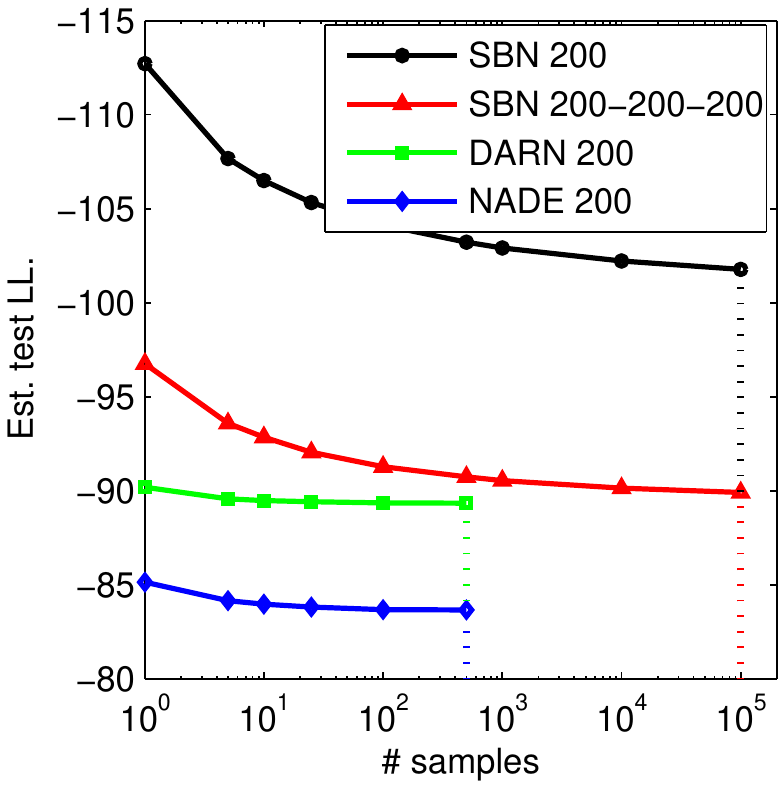}
}
\end{center}\vspace{-.5cm}
\caption{Log-likelihood estimation on \mnist{} w.r.t
(Left) number of samples $S$ used during training and
(Right) number of samples $K$ used during estimating the log-likelihood.
Dotted line marks the results estimated with $S=5$ and $K=10^5$ for SBN and $K=500$ for \darn{} and NADE
, as reported in \tabl{table:logpvs-methods}.}
\vskip -.3cm
\end{figure}

\vspace{-.1cm}
\subsection{Discrete Hidden Variable Models}\label{sec:exp:discrete}

\vspace{-.1cm}
\subsubsection{Binarized \mnist{}}
\vspace{-.1cm}

\begin{figure*}
\vspace{-.2cm}
\begin{center}
\subfigure{\label{fig:visualization:data}
\includegraphics[width=0.245\textwidth]{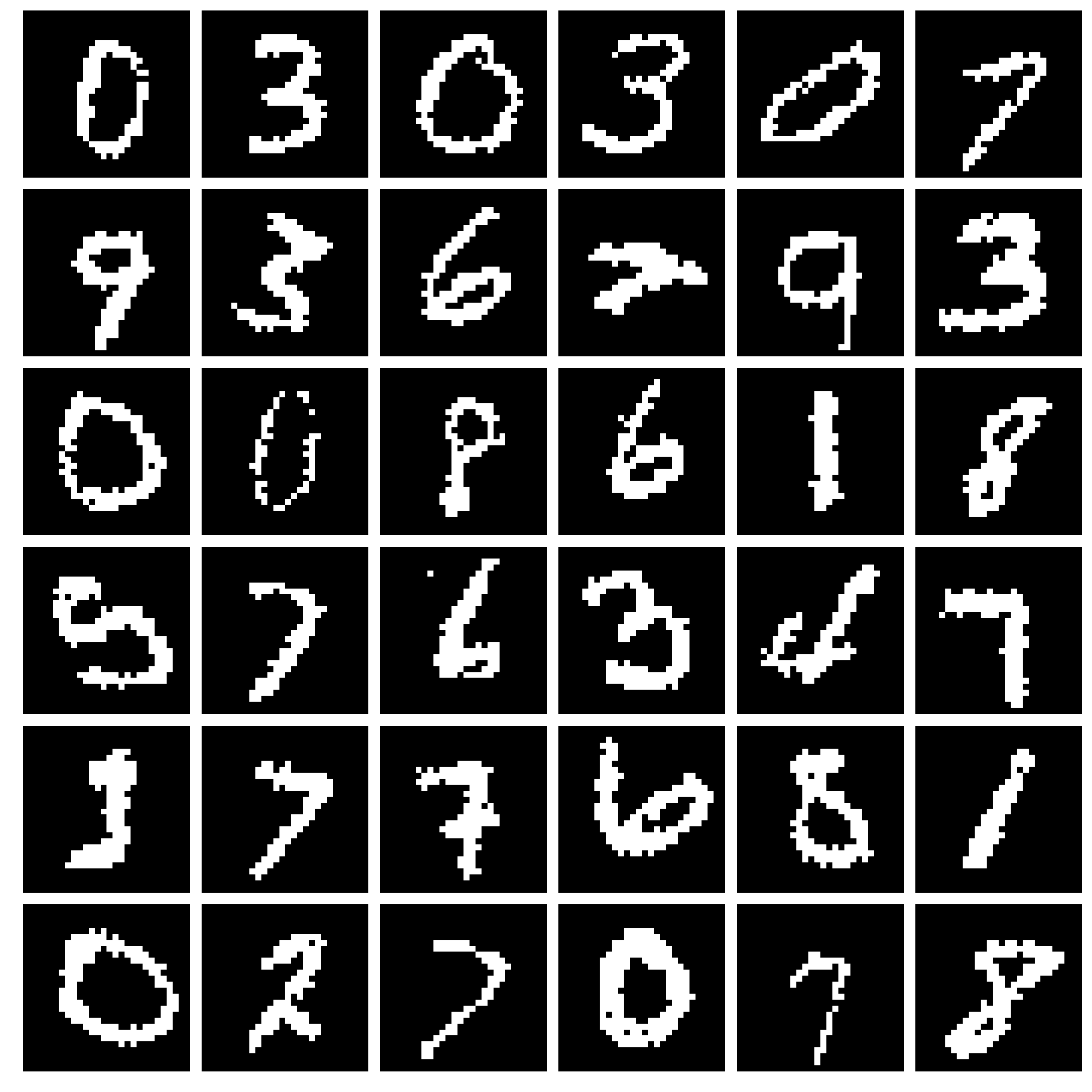}
}
\hspace{-.3cm}
\subfigure{\label{fig:visualization:sample}
\includegraphics[width=0.245\textwidth]{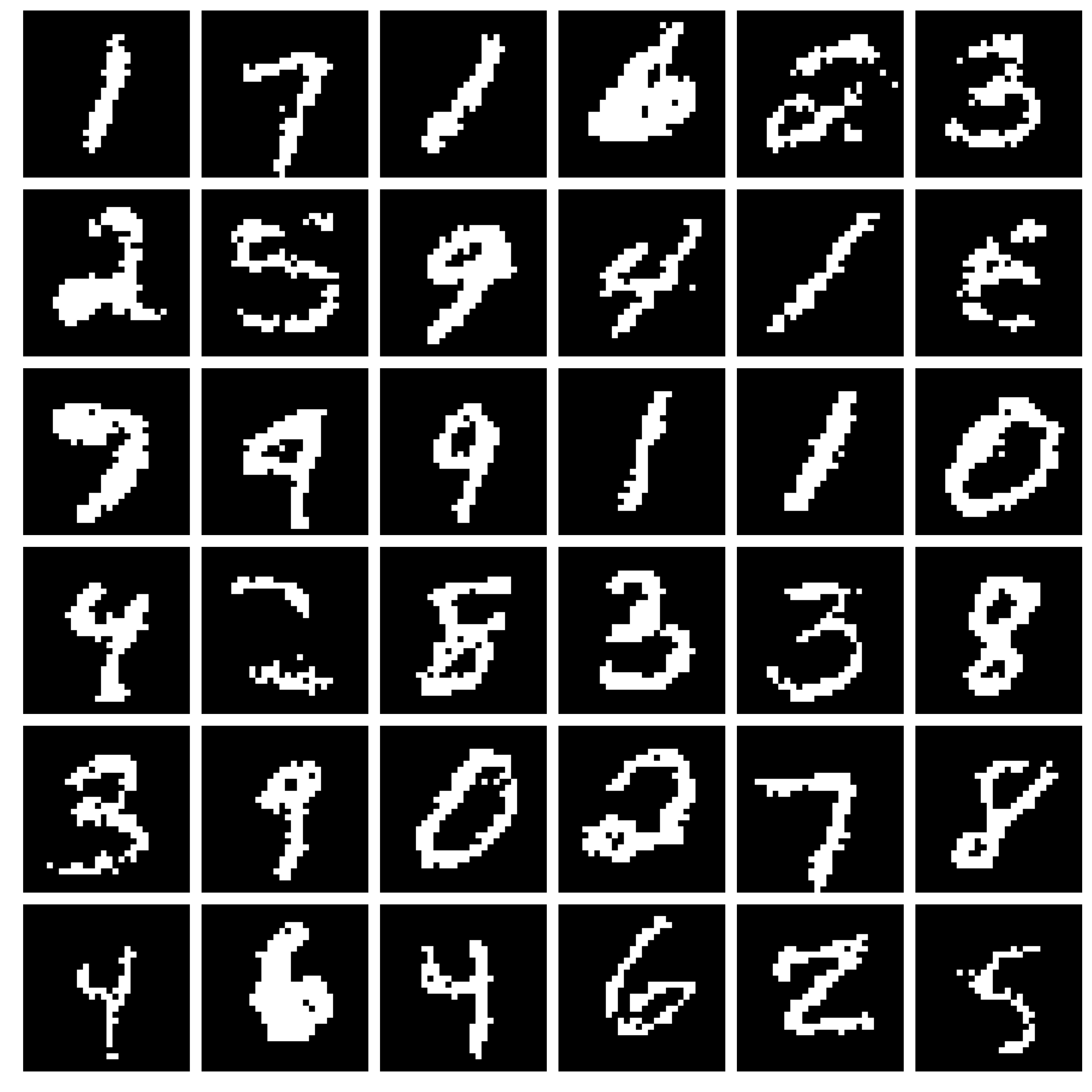}
}
\hspace{-.3cm}
\subfigure{\label{fig:visualization:feature-sparse}
\includegraphics[width=0.245\textwidth]{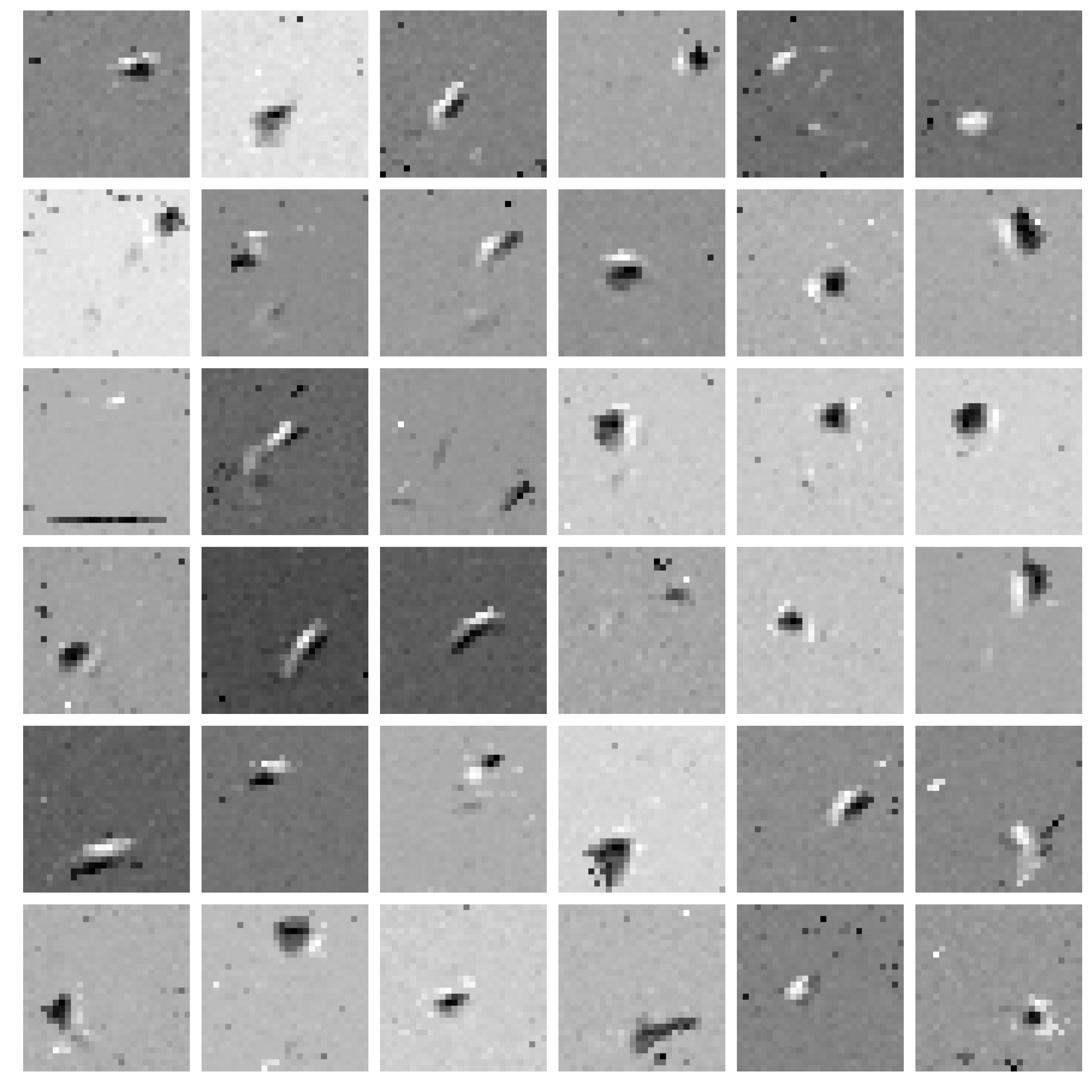}
}
\hspace{-.3cm}
\subfigure{\label{fig:visualization:feature-normal}
\includegraphics[width=0.245\textwidth]{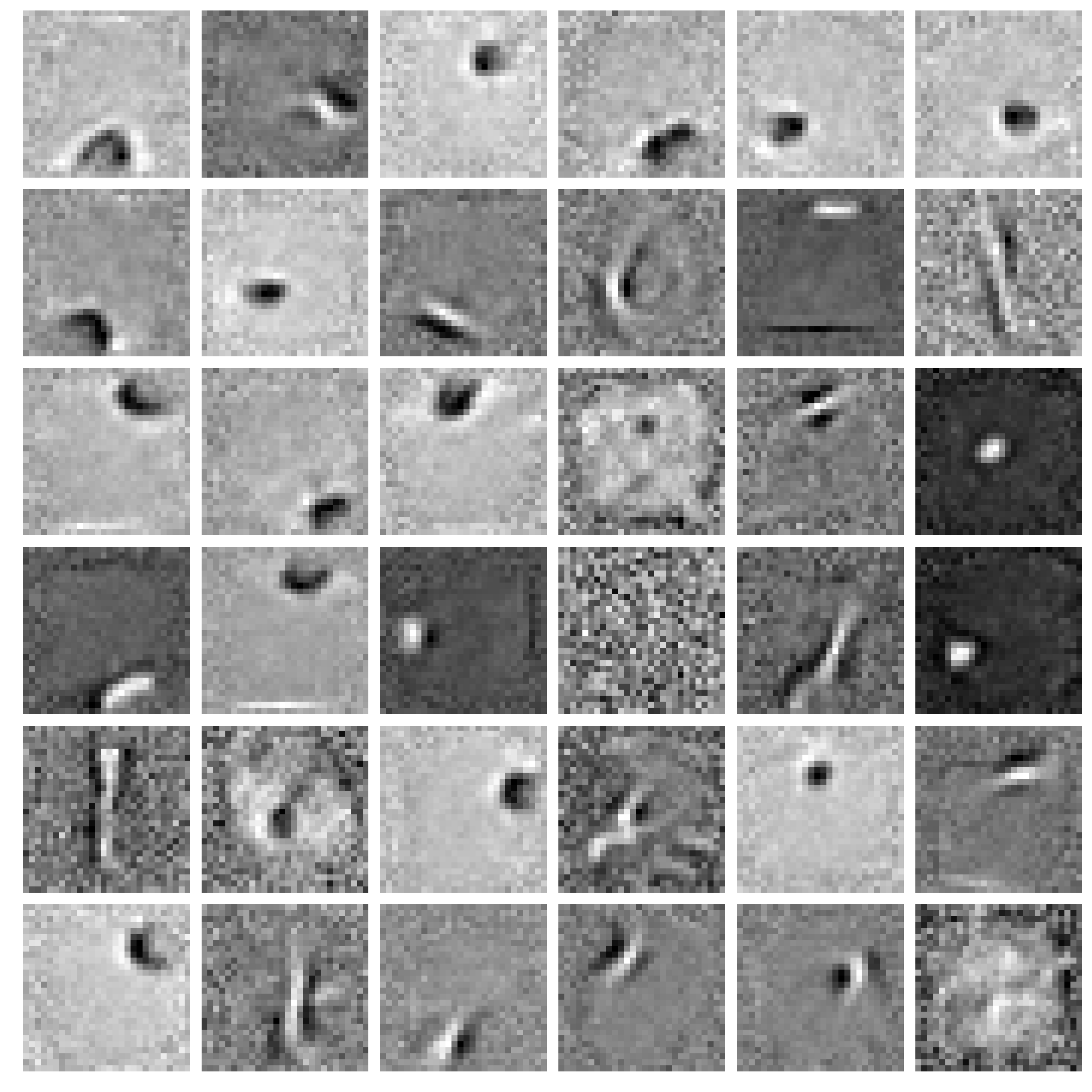}
}\\
\vspace{-.3cm}

\subfigure[(a)]{
\includegraphics[width=0.245\textwidth]{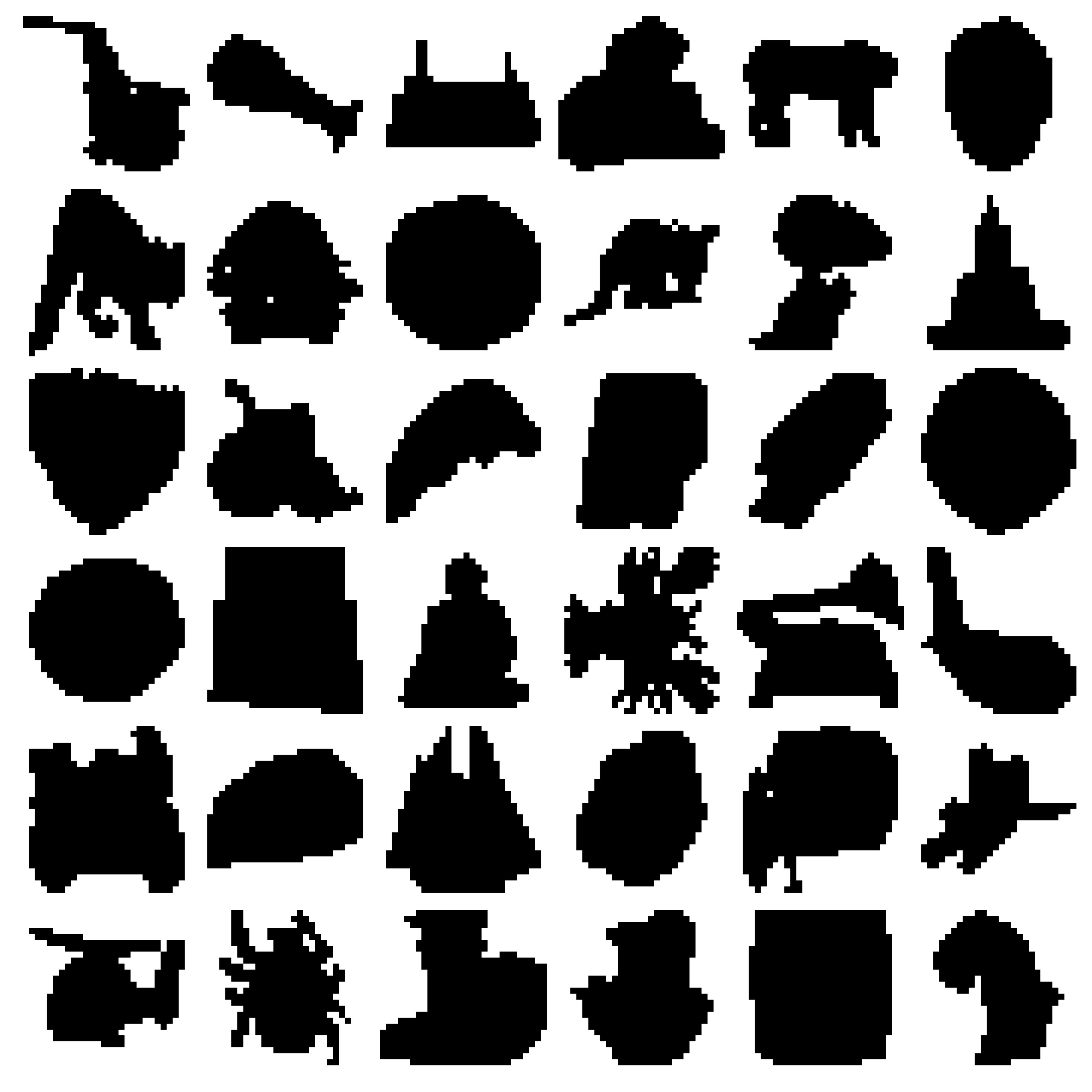}
}
\hspace{-.3cm}
\subfigure[(b)]{
\includegraphics[width=0.245\textwidth]{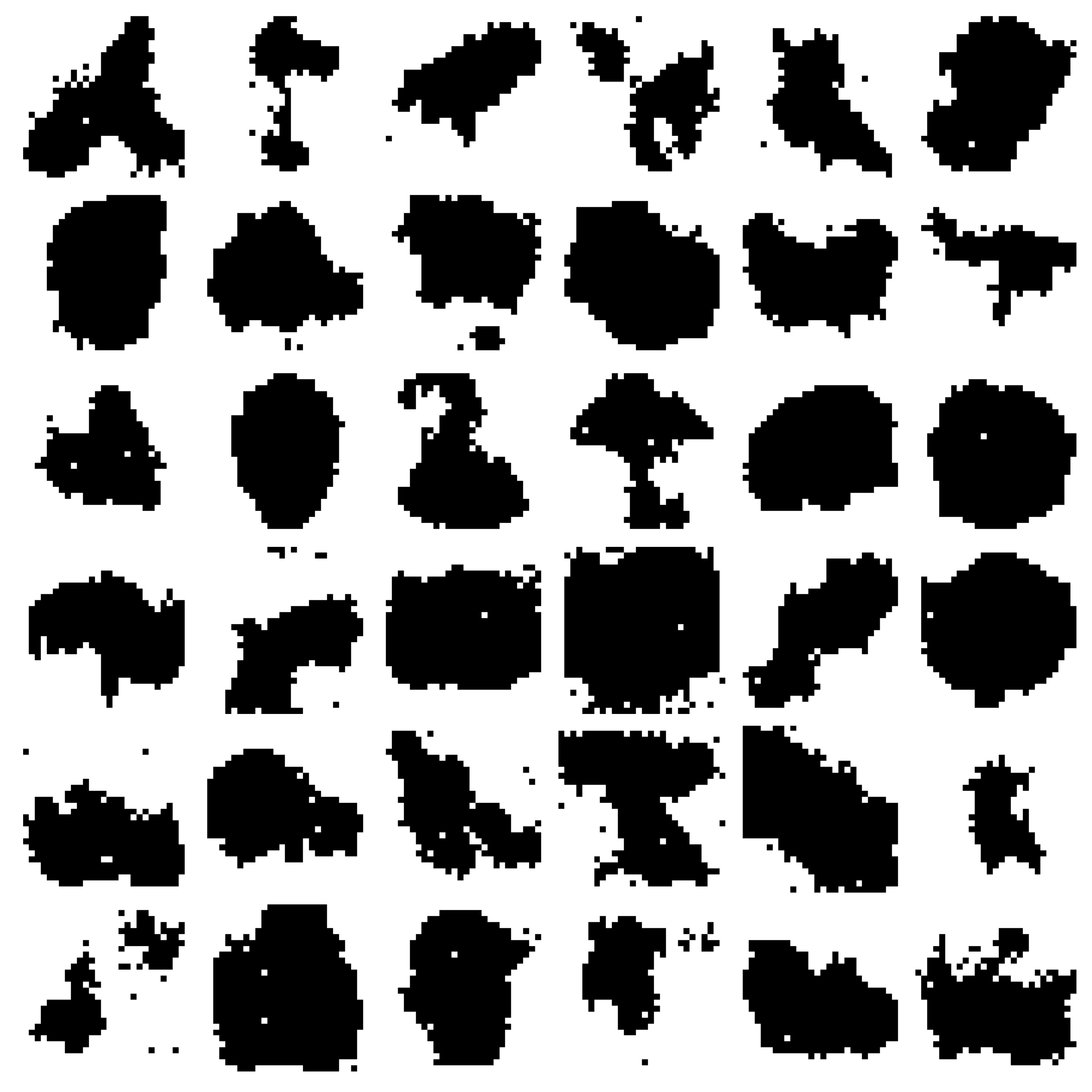}
}
\hspace{-.3cm}
\subfigure[(c)]{
\includegraphics[width=0.245\textwidth]{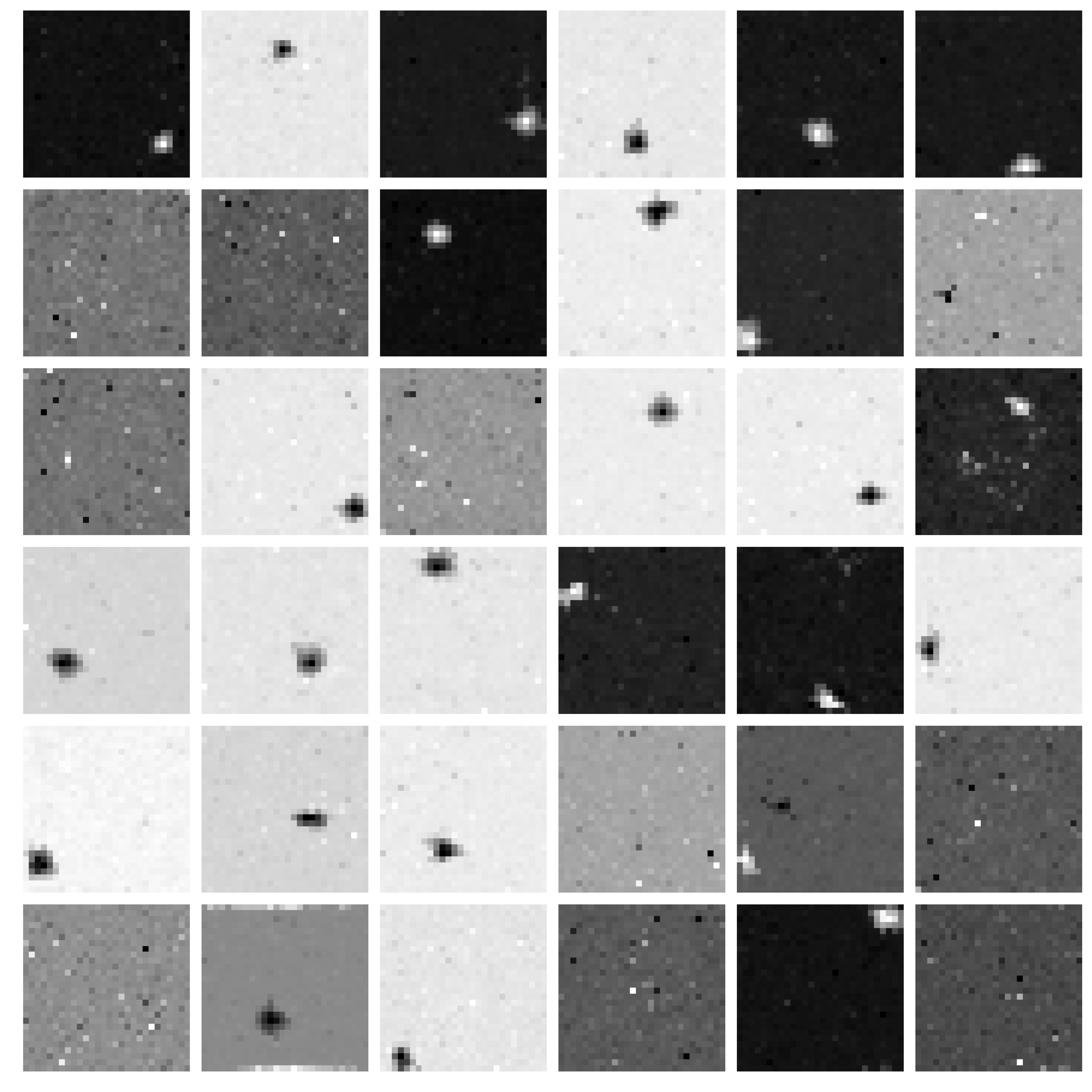}
}
\hspace{-.3cm}
\subfigure[(d)]{
\includegraphics[width=0.245\textwidth]{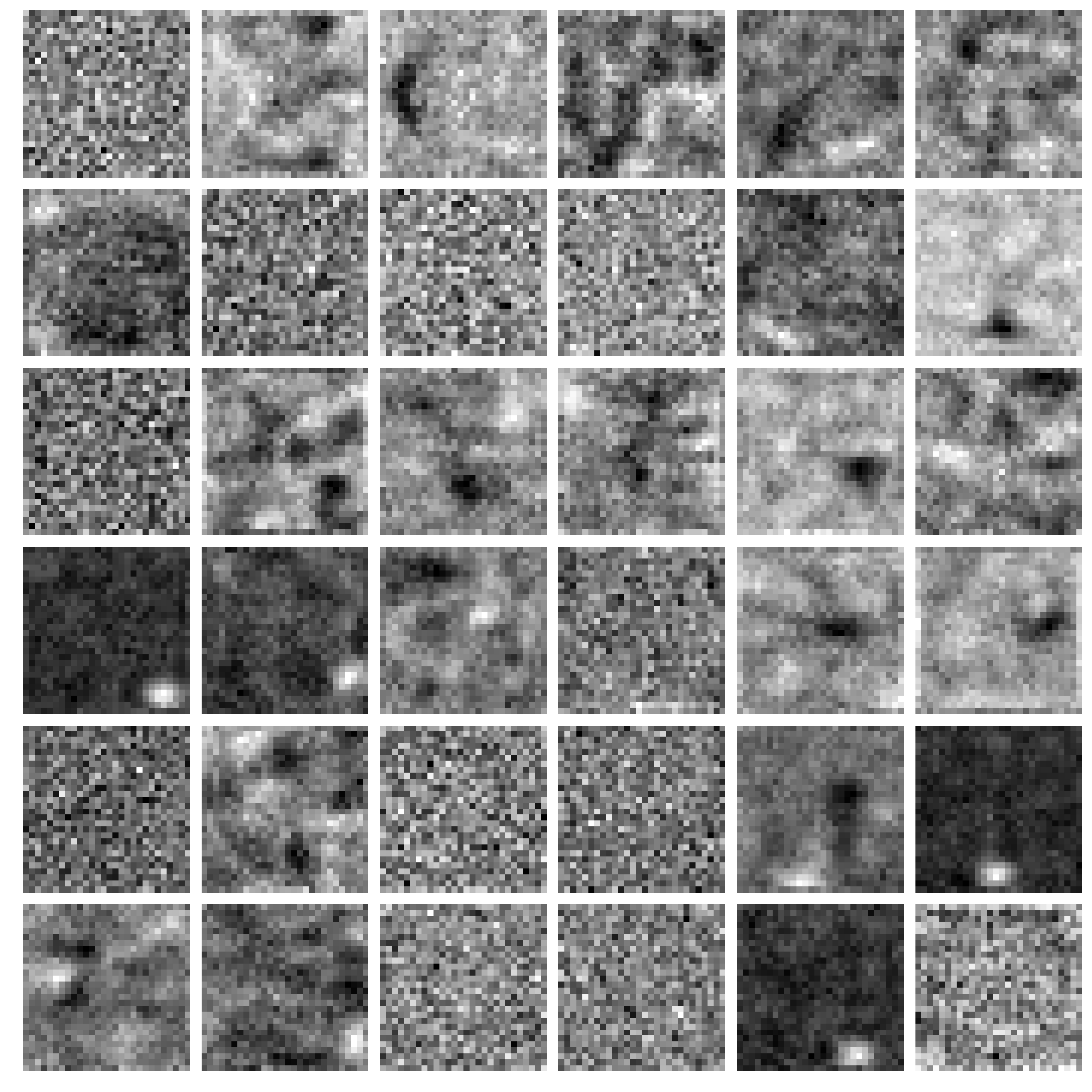}
}
\end{center}
\vspace{-.4cm}
\caption{Visualization: (a) Training data. (b) Samples from the learned models: NADE/NADE 200 for \mnist{} and NADE/NADE 150 for \caltech{}. (c) Features at the bottom layer learned with sparse prior. (d) Features at the bottom layer learned with normal prior. (Top) \mnist{}. (Bottom) \caltech{}.}
\label{fig:visualization}
\vskip -.2cm
\end{figure*}

The binarized \mnist{} dataset consists of $50,000$ training samples, $10,000$ validation samples and $10,000$ test samples.

We consider five benchmark models: three SBN models, one \darn{} model and one NADE model.
For the three models with SBN layers, we also use SBN layers to construct the \recmodel{};
for the two models with \darn{} layer and NADE layer, we follow \citet{bornschein2014reweighted} to use NADE layer for the \recmodel{}.
The model sizes and the results are summarized in \tabl{table:logpvs-methods}. Details of the construction of the models are summarized in \splm{}.

\begin{table}[t]\vspace{-.1cm}
\caption{\mnist{} results of various methods and models. Results are taken from [1] \citet{bornschein2014reweighted},
[2] \citet{Larochelle:11}, [3] \citet{Uria2013b}, [4] \citet{gregor14deep}, [5] \citet{Salakhutdinov08on},
[6] \citet{raiko2014iterative}, [7] \citet{murray2009evaluating}, [8] \citet{burdaiwae}. $\diamond$ Trained on the original \mnist{} dataset.}
\label{table:logpvs-mnist}
\vspace{-.1cm}
\begin{center}
\begin{small}
\begin{tabular}{lccc}
\hline
Methods (Models)                   &  &  & Est. test LL. \\
\hline
\methodq{} (NADE/NADE 200)         &  &  & $\mathbf{-83.67}$ \\
\hline
RWS (NADE/NADE 250) [1]            &  &  & $-85.23$ \\
RWS (ARSBN/SBN 500) [1]            &  &  & $-84.18$ \\
NADE (500 hidden units) [2]        &  &  & $-88.86$ \\
EoNADE 2hl (128 orderings) [3]     &  &  & $-85.10$ \\
DARN (500 hidden units) [4]        &  &  & $-84.13$ \\
RBM  (500 hidden units) [5]        &  &  & $-86.34$ \\
EoNADE-5 2HL(128 Ords) [6]         &  &  & $-84.68$ \\
DBN 2hl [7] $^\diamond$            &  &  & $-84.55$ \\
IWAE 2sl [8]                       &  &  & $-85.32$ \\
\hline
\end{tabular}
\end{small}
\end{center}
\vspace{-0.6cm}
\end{table}

We first investigate the effect of the neural adaptive importance sampler (NAIS).
For comparison, we also estimate the gradient \eqn{eq:derivative} by directly sampling from $\pz$ using a Gibbs sampler.
(The derivation for the Gibbs sampler is included in \splm{}.)
We denote the resulting method by \methodg{}.
In \tabl{table:logpvs-methods} we observe that the DSGNHT using a NAIS consistently outperforms the DSGNHT using a Gibbs sampler,
especially for deeper models and autoregressive models,
since the model parameters are higher-dimensional and highly correlated.
We then compare our method to several other state-of-the-art methods on the five benchmark models.
We observe that our method outperforms RWS almost on all models, except for \darn{} 200, on which we are slightly worse than RWS.

\begin{figure}[tb]\vspace{-.2cm}
\begin{center}
\subfigure{
\includegraphics[width=0.44\textwidth]{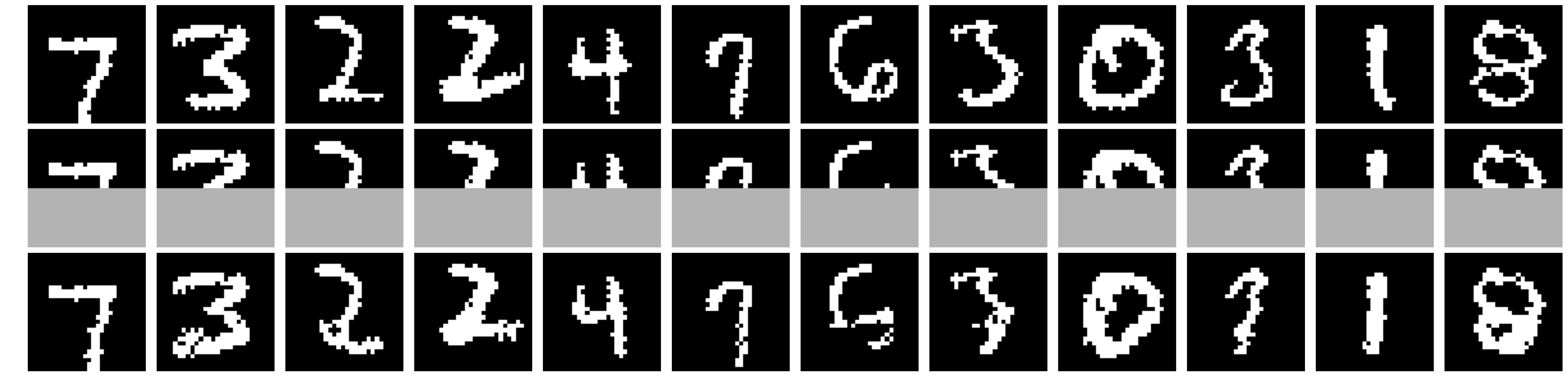}
}\\
\vspace{-.3cm}
\subfigure{
\includegraphics[width=0.44\textwidth]{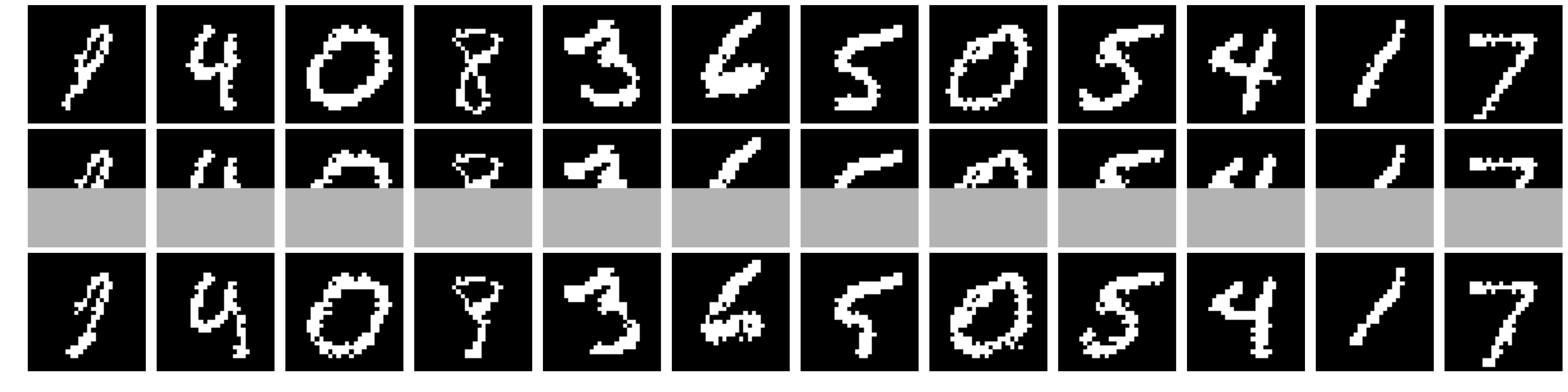}
}
\end{center}\vspace{-.6cm}
\caption{\mnist{} missing data prediction by SBN 200-200-200:
(Top) Original data. (Middle) Hollowed data. (Bottom) Reconstructed data.}
\label{fig:visualization:miss}
\vskip -0.3in
\end{figure}

We compare our best result to the state-of-the-art results on binarized \mnist{} in \tabl{table:logpvs-mnist}.
The NADE/NADE 200 model achieves an \estll{} of $-83.67$, which outperforms most published results.
\citet{gregor2015draw} give a lower bound $-80.97$, which exploits spatial structure.
IWAE \cite{burdaiwae} achieves $-82.90$, which is trained on the original \mnist{} dataset \cite{lecun1998gradient} and thus not directly comparable.
We cite their results on the binarized \mnist{} in \tabl{table:logpvs-mnist}.

In \fig{fig:post-samples}, we investigate the influence of the number of samples $M$ on the posterior mean estimator \eqn{eq:posterior-mean}.
We can observe that on all models using more samples for posterior mean brings consistent improvements.
On SBN/SBN models using $M=100$ samples improves around $0.6$ nat than using $M=1$ (in which case one posterior sample is estimated).
On autoregressive models, using $M=100$ samples brings an improvement more than $2$ nats.

We show the influence of the number of samples $S$ used during training in \fig{fig:train-samples}.
we observe that the \estll{} on test data improves as $S$ grows up.
\fig{fig:test-samples} presents the curves of the final estimated test log-likelihood with respect to the number of samples $K$ used for the estimator \eqn{eq:estimation}. We observe that $K=100,000$ and $K=500$ are large enough to get a good \estll{} for SBN and \darn{}(NADE) models, respectively.

\fig{fig:visualization} visualizes the generative performance of the learned models.
In \fig{fig:visualization:data}, we show the randomly sampled training data of \mnist{} and \caltech{}.
\fig{fig:visualization:sample} displays the examples generated from the learned models.
We observe that the generated samples are visually good.

One advantage of Bayesian framework is that we can specify sparsity-encouraging priors on the model parameters explicitly,
e.g., the Student-t prior in our experiments. \fig{fig:visualization:feature-sparse} and \fig{fig:visualization:feature-normal}
demonstrate the difference between features learned with a sparse (Student-t prior) prior and a non-sparse (Gaussian) prior.
We observe that the features learned with a sparse prior appear more localized.

We further demonstrate the ability of the learned models on predicting missing data.
For each test image, the lower half is assumed missing and the upper half is used to inference the hidden units \cite{Gan:aistats15}.
Then, with the hidden units, the lower half is reconstructed.
Prediction is done by repeating this procedure and finally sampling from the generative model with the inferred hidden units.
\fig{fig:visualization:miss} demonstrates some example completions for the missing data on MNIST.

\begin{table}[t]\vspace{-.3cm}
\caption{\caltech{} results of various training methods and models. Results are taken from [1] \citet{bornschein2014reweighted}, [2] \citet{cho2013enhanced},
[3] \citet{raiko2014iterative}. $\dagger$ Results are produced using the authors' published code.}
\label{table:logpvs-caltech}
\vspace{-.1cm}
\begin{center}
\begin{small}
\begin{tabular}{lc}
\hline
Methods (Models)                     & Est. test LL. \\
\hline
\methodq{} (SBN/SBN 200)             & $-122.7$ \\
\methodq{} (SBN/SBN 200-200)         & $-108.0$ \\
RWS (SBN/SBN 200)$\dagger$           & $-134.4$ \\
RWS (SBN/SBN 200-200)$\dagger$       & $-126.0$ \\
\hline
\hline
\methodq{} (SBN/SBN 200-200-200)     & $-105.2$ \\
\methodq{} (SBN/SBN 300-100-50-10)   & $\mathbf{-103.6}$ \\
\methodq{} (NADE/NADE 150)           & $\mathbf{-100.0}$ \\
\hline
RWS (SBN/SBN 300-100-50-10) [1]      & $-113.3$ \\
RWS (NADE/NADE 150) [1]              & $-104.3$ \\
NADE (500 hidden units) [1]          & $-110.6$ \\
RBM (4000 hidden units) [2]          & $-107.8$ \\
NADE-5 (4000 hidden units) [3]       & $-107.3$ \\
\hline
\end{tabular}
\end{small}
\end{center}
\vspace{-0.6cm}
\end{table}

\vspace{-.1cm}
\subsubsection{\caltech{}}
\vspace{-.1cm}

The \caltech{} dataset consists of $4,100$ training samples, $2,264$ validation samples and $2,307$ test samples.
We first compare our method to RWS on two benchmark models in \tabl{table:logpvs-caltech} (Top)
and observe that our method achieves significant improvements. On SBN/SBN 200-200, we get a test \estll{} of $-108.0$
which improves over RWS for $18$ nats.

\tabl{table:logpvs-caltech} (Bottom) summarizes ours best results and other state-of-the-art results.
Our NADE/NADE 150 network reaches a test \estll{} of $-100.0$, which improves RWS on the same model for $4.3$ nats.
We observe a remarkable effect of increasing the number of samples $M$ for posterior mean:
the test \estll{} of $-100.0$ at $M=100$ improves $5.3$ nats compared to $-105.3$ at $M=1$.
\citet{Gan:aistats15} achieve $-96.40$ by training FVSBN \cite{frey1998graphical} with both training data and validation data.
A latest work by \citet{goessling2015sparse} achieves $-88.48$ by developing a mixture model of sparse autoregressive network.
\fig{fig:visualization:sample} visualizes the samples drawn from the learned models.

\begin{table}[tb]\vspace{-.3cm}
\caption{Results of log-likelihood estimation on single-stochastic-layer variational auto-encoder. Results of VAE and IWAE are taken from \citet{burdaiwae}.}
\label{table:logpvs-vae}
\vspace{-.5cm}
\begin{center}
\begin{small}
\begin{tabular}{lccc}
\hline
Dataset                        &   VAE     &   IWAE    &  \methodq{} \\
\hline
binarized \mnist{}             & $-88.83$  & $-87.63$  & $\mathbf{-86.93} $   \\
\omni{}                        & $-107.62$ & $-106.12$ & $\mathbf{-106.10}$   \\
\hline
\end{tabular}\vspace{-.6cm}
\end{small}
\end{center}

\end{table}

\vspace{-.1cm}
\subsection{Variational Auto-Encoders}\label{sec:exp:continuous}
\vspace{-.1cm}

Finally, we consider the DGMs with continuous hidden variables.
One popular example is the variational auto-encoder (VAE)~\cite{kingma14iclr}.
Intuitively, the posterior of DGMs with continuous hidden variables is harder to capture,
as the hidden variables have much more freedom compared to that of DGMs with discrete hidden variables.
Such freedom potentially results in high variance of the gradients estimation.
VAE and the importance weighted auto-encoders (IWAE) \cite{burdaiwae} alleviate this problem by adopting a reparametrization trick.
Then the variational parameters $\phi$ can be optimized tying to the generative model.

In our \methodq{}, we indeed observe high variance of the gradients estimation.
However in theory, any distribution that satisfies $\qz > 0$ wherever $\pz > 0$ can be used as a proposal.
Such a property makes any other reasonable objectives (instead of the inclusive KL-divergence described in \secref{sec:dsmcmc:dsmcmc}) for $\qz$ is adoptable.
We adopt the same objective in IWAE for the proposal distribution
\footnote{We use the objective in IWAE for optimizing the proposal distribution $\qz$ only, leaving other part of our method unchanged.}
and find it works well in practice.

We follow IWAE to train a single-stochastic-layer VAE with 50 hidden units.
In between the data and the hidden variables are two deterministic layers with tanh activation.
The model is trained on the binarized \mnist{} and the \omni{} datasets.
We use the \omni{} dataset downloaded from \citet{burdaiwae} which
consists of $23,000$ training samples, $1,345$ validation samples and $87,00$ test samples.
We use $M=200$ for the posterior mean estimator
and follow IWAE to use $K=5,000$ to evaluate the test \estll{}.
Since IWAE also adopts an importance sampler to estimate the objective (as well as its gradient),
we compare the results using $S=5$ samples during training for both IWAE and our method.
We achieve comparable or better results as summarized in \tabl{table:logpvs-vae}.

\vspace{-.1cm}
\section{Conclusions and Future Work}
\vspace{-.1cm}

We propose a powerful Bayesian inference method based on stochastic gradient MCMC for deep generative models with continuous parameter space.
It enjoys several advantages of Bayesian formalism such as sparse Bayesian inference.
Our results include state-of-the-art performance on standard published datasets.

For future work we like to investigate the performance on learning sparse Bayesian models. Also, learning nonparametric Bayesian DGMs is another interesting challenge.



\bibliography{bibtex}
\bibliographystyle{icml2016}

\appendix

\section{Model Setup}

We describe how the models are constructed with the conditional stochastic layers.
Each model should consists of a top layer $p(\zv^{(L)}|\thetav)$,
a data layer $p(\xv|\zv^{(1)},\thetav)$ and (optionally) several intermediate layer $p(\zv^{(l)}|\zv^{(1+1)},\thetav)$.
The output of the layer $(l+1)$ (random samples) is passed as the input to the layer $l$
and thus a full generative process for the data is built.
In principle, the type of each layer can be chosen arbitrarily,
as long as the input dimension and the output dimension of adjacent layers match to each other.

The \recmodel{} can be constructed in a similar way.
Given a generative model with $L$ hidden layers,
the \recmodel{} should also contains $L$ stochastic layers.
The first layer takes $x$ as input and outputs random samples of $\zv^{(1)}$.
The output of the layer $l$ (random samples of $\zv^{(l)}$) is passed as the input to the layer $l+1$.
Note the type of each layer can also be chosen arbitrarily,
as long as the dimensions of adjacent layers match to each other and the output dimension of the $l$-th layer matches the input dimension of the $l$-th layer of the generative model.

In our experiments, all tested models and their recognition models consist only one type of stochastic layer.
In the following we describe the detailed architectures of our tested models.

\textbf{SBN/SBN models:}

For all models with SBN layers we construct the \recmodel{} with SBN layers too.
We use four SBN/SBN architectures in the experiments:
1-hidden-layer with 200 hidden units (SBN/SBN 200);
2-hidden-layer with 200 hidden units in each hidden layer (SBN/SBN 200-200);
3-hidden-layer with 200 hidden units in each hidden layer (SBN/SBN 200-200-200);
and 4-hidden-layer with 300(closest to data), 100, 50, 10 hidden units (SBN/SBN 300-100-50-10).

We use the following top layer (equivalent to factorized Bernoulli distribution):
\begin{equation}\label{eq:prob-sbn-top}
p(z^{(L)}_i = 1|\thetav) = \sigma(b_i),
\end{equation}
and the likelihood model:
\begin{equation}\label{eq:prob-sbn-like}
p(z^{(l)}_i = 1|\zv^{(l+1)},\thetav) = \sigma(\Wv^{(l)}_{i,:}\zv^{(l+1)}+b_i^{(l)}),
\end{equation}
where we define $\zv^{(0)}=\xv$.
The model parameters are $\thetav=\{\Wv^{(l)},\bv^{(l)}\}_{l=0}^{L-1}\cup\{\bv^{(L)}\}$.

\textbf{DARN/NADE models:}

we follow \citet{bornschein2014reweighted} to use NADE layers in the \recmodel{} for the \darn{} models.
We test a shallow model (1-hidden-layer \darn{}/NADE 200) in our experiments.
We use the following top layer (equivalent to FVSBN~\cite{frey1998graphical}):
\begin{equation}\label{eq:prob-darn-top}
p(z^{(L)}_i = 1|\zv^{(L)}_{<i}, \thetav) = \sigma(\Wv_{i,<i}^{(L)}\zv^{(L)}_{<i}+b_i^{(L)}),
\end{equation}
and the likelihood model:
\begin{equation}\label{eq:prob-darn-like}
\begin{aligned}
p(z^{(l)}_i &= 1|\zv^{(l)}_{<i}, \zv^{(l+1)}, \thetav) = \\
& \sigma(\Uv_{i,:}^{(l)}\zv^{(l+1)}+\Wv_{i,<i}^{(l)}\zv^{(l)}_{<i}+b_i^{(l)}).
\end{aligned}
\end{equation}
The model parameters are $\thetav=\{\Uv^{(l)}\}_{l=0}^{L-1}\cup\{\Wv^{(l)},\bv^{(l)}\}_{l=0}^{L}$.

\textbf{NADE/NADE models:}

We test a shallow model (1-hidden-layer NADE/NADE 200) in our experiments.
We use the following top layer:
\begin{equation}\label{eq:prob-nade-top}
\begin{aligned}
p(z^{(L)}_i = 1|\zv^{(L)}_{<i}, \thetav) = &  \\
\sigma\Big(\Vv_{i,:}^{(L)}\sigma(\Wv_{:,<i}^{(L)}&\zv^{(L)}_{<i}+\av^{(L)})+b_i^{(L)}\Big),
\end{aligned}
\end{equation}
and the likelihood model:
\begin{align}\label{eq:prob-nade-like}
p(z^{(l)}_i& = 1| \zv^{(l)}_{<i}, \zv^{(l+1)}, \thetav) = \sigma\Big(\Vv_{i,:}^{(l)}\sigma(\Wv_{:,<i}^{(l)}\zv^{(l)}_{<i} \nonumber \\
+&\Uv^{(l)}\zv^{(l+1)}+\av^{(l)})+\Rv^{(l)}_{i,:}\zv^{(l+1)}+b_i^{(l)}\Big),
\end{align}
The model parameters are $\thetav=\{\Uv^{(l)},\Rv^{(l)} \}_{l=0}^{L-1}\cup\{\Vv^{(l)},\Wv^{(l)},\av^{(l)},\bv^{(l)} \}_{l=0}^{L}$.

\begin{algorithm*}[!t]
   \caption{A Detailed Version of Doubly Stochastic Gradient MCMC with Neural Adaptive Proposals}
   \label{algo:dsmcmc-detail}
\begin{algorithmic}
   \STATE {\bfseries Input:} $\Xv=\{\xv_1,\cdots,\xv_N\}$: the dataset
   \STATE {\bfseries Input:} $S$: number of samples used during training
   \STATE {\bfseries Input:} $M$: number of samples used for computing the posterior mean estimation \eqn{eq:posterior-mean}
   \STATE {\bfseries Input:} $\gamma$: per-batch learning rate for SGNHT, $|B|$: mini-batch size, $a$: momentum decay
   \STATE {\bfseries Input:} $\eta'$: step size for Adam
   \STATE {\bfseries Input:} $n_{\thetav}$, $n_{\phiv}$: number of $\thetav$ or $\phiv$ update during each mini-batch
   \STATE Initialize $\thetav$, $\phiv$: following the heuristic of~\citet{glorot2010understanding}
   \STATE Initialize $\uv\sim \N(\zerov,\eta\Iv)$, $\alphav=a\Iv$
   \FOR{epoch $=1,2,\cdots$}
   \STATE Randomly split the data $\Xv$ into mini-batches $B_1,\cdots,B_{N/|B|}$
   \FOR{mini-batch $B_i$ in $\{B_1,\cdots,B_{N/|B|}\}$}
   \FOR{$i_{\thetav} = 1,\cdots,n_{\thetav}$}
   \STATE Sample $\{\zv_n^{(s)}\}_{s=1}^S$ from the proposal $q(\zv|\xvn; \phiv)$, $n \in B_i$ 
   \STATE Estimate the gradients $\nabla \log p(\xvn | \thetav)$ with \eqn{eq:gradient-selfnormalize}, $n \in B_i$
   \STATE Update $\thetav$ with \eqn{eq:sgnht}
   \FOR{$i_{\phiv} = 1,\cdots,n_{\phiv}$}
   \STATE Sample $\{\zv_n^{(s)}\}_{s=1}^S$ from the proposal $q(\zv|\xvn; \phiv)$, $n \in B_i$
   \STATE Estimate the gradients $\nabla_{\phiv} \mathcal{J}(\phiv;\thetav,\Xv)$ with \eqn{eq:gradient-recmodel-batch}, $n \in B_i$
   \STATE Update $\phiv$ using Adam optimizer with $\nabla_{\phiv} \mathcal{J}(\phiv;\thetav,\Xv)$
   \ENDFOR
   \ENDFOR
   \ENDFOR
   \ENDFOR
   \STATE Run another $M$ epochs to estimate the posterior mean $\hat{\thetav}$
   \STATE {\bfseries Output:} $\hat{\thetav}$
\end{algorithmic}
\end{algorithm*}

\textbf{VAE/VAE models:}

For the VAE model, we follow \citet{kingma14iclr} to use an isotropic multivariate Gaussian top layer:
\begin{equation}\label{eq:prob-vae-top}
p(z^{(L)}_i = 1|\thetav) = \N(\zerov,\Iv).
\end{equation}
The VAE stochastic layer itself contains an internal MLP.
In our experiments, we train single-stochastic-layer VAE with 50 hidden units.
In between the data and the hidden variables are two deterministic layers with tanh activation.
The dimension of the two deterministic layers are both 100.
The \recmodel{} is a stochastic VAE layer within which are two 100-dimensional deterministic layers.
Such an architecture is used in \citet{burdaiwae}.

In experiments of training VAE, we adopt the objective in IWAE~\cite{burdaiwae} for learning the proposal distribution $\qz$:
\begin{equation}
\mathcal{L}_K = \ep_{\zv^{(k)}\sim\qz}\left[\log\frac{1}{K}\sum_{k=1}^{K}\frac{p(\xv,\zv^{(k)})}{q(\zv^{(k)}|\xv)}\right].
\end{equation}
Then the gradient can be evaluated by adopting the reparametrization trick~\cite{kingma14iclr}:
\begin{align*}
\nabla_{\phiv}\mathcal{L}_K=& \nabla_{\phiv}\ep_{\zv^{(k)}\sim\qz}\left[\log\frac{1}{K}\sum\nolimits_{k=1}^{K}\frac{p(\xv,\zv^{(k)})}{q(\zv^{(k)}|\xv)}\right] \\
=& \nabla_{\phiv}\ep_{\epsilonv^{(k)}\sim p(\epsilonv)}\left[\log\frac{1}{K}\sum\nolimits_{k=1}^{K}w(\xv,\zv(\epsilonv^{(k)},\xv,\phiv))\right] \\
=& \ep_{\epsilonv^{(k)}\sim p(\epsilonv)}\left[\nabla_{\phiv}\log\frac{1}{K}\sum\nolimits_{k=1}^{K}w(\xv,\zv(\epsilonv^{(k)},\xv,\phiv))\right] \\
=& \ep_{\epsilonv^{(k)}\sim p(\epsilonv)}\left[\sum\nolimits_{k=1}^{K}\widetilde{w_k}\nabla_{\phiv}\log w(\xv,\zv(\epsilonv^{(k)},\xv,\phiv))\right], \\
\end{align*}\\[-1cm]
where we have omitted the model parameters $\thetav$ in the above gradients, since $\thetav$ is fixed when learning the proposal distribution. In the above derivations, $\epsilonv^{(1)},\cdots,\epsilonv^{(K)}$ are the auxiliary variables as defined in VAE.
$w_k=w(\xv,\zv(\epsilonv^{(k)},\xv,\phiv))=\frac{p(\xv,\zv(\epsilonv^{(k)},\xv,\phiv))}{q(\zv(\epsilonv^{(k)},\xv,\phiv)|\xv)}$ are the importance weight and $\widetilde{w_i}$ are the normalized importance weights
as defined in IWAE.

\section{Experimental Setup}

We describe our experimental setup here, including the parameter setting and implementation details.

In our implementation, we use the reformulated form of multivariate SGNHT~\cite{ding2014bayesian}:
\begin{align}
\thetav_{t+1} &= \thetav_{t} + \uv_{t}, \label{eq:sgnht} \\
\uv_{t+1} &= \uv_{t} - \xiv_{t}\odot\uv_{t}-\eta\nabla_{\thetav} \tilde{U}(\thetav_{t+1}) + \N(\zerov,2a\eta\Iv), \nonumber \\
\alphav_{t+1} &= \alphav_{t} + (\pv_{t+1}\odot\pv_{t+1}-\eta\Iv), \nonumber
\end{align}
where we have setting $\uv=\lambda\pv$, $\eta=\lambda^2$, $\alphav=\lambda\xiv$ and $a=A\lambda$.
This reformulation is cleaner and easier to implement.
In analog to SGD with momentum, $\eta$ is called the learning rate and $\mathbf{1}-\alphav$ are the momentum terms~\cite{Chen:icml14}.
The initialization of SGNHT is as follows:
$\uv$ is random sampled from $\N(\zerov,\eta\Iv)$ and
$\alphav$ is initialized as $a\Iv$.
There are three parameters for SGNHT:
the learning rate $\eta$,
the momentum decay $a$,
and the mini-batch size $B$.
In our implementation we choose the mini-batch size $B=100$, the momentum decay $a=\{0.1,0.01\}$.
For numerical stability, we choose $\eta=\frac{\gamma}{N}$, where $\gamma$ is called the ``per-batch learning rate''~\cite{Chen:icml14}.
The per-batch learning rate $\gamma$ is chosen from $ \{0.01, 0.005, 0.001\}$
with best performance.

For the \recmodel{}, we use Adam~\cite{kingma15adam} to learn the parameters $\phiv$. There are four parameters for Adam:
the stepsize $\eta'$, the exponential decay rates $\{\beta_1$, $\beta_2\}$ and $\epsilon$ which is used to prevent division by zero.
In our implementation we choose $\beta_1 = 0.9$, $\beta_2 = 0.999$ and $\epsilon=10^{-10}$. The stepsize $\eta'$ is chosen from $\{1,3,5\}\times 10^{-4}$ with best performance.

The model parameters are initialized following the heuristic of~\citet{glorot2010understanding}.
The Student-t's prior for all model parameters are set with a scale parameter $\sigma = 0.09$, location parameter $\mu = 0$ and degrees of freedom $\nu = 2.2$.

Our method involves updating the generative model parameters $\thetav$ and the \recmodel{} parameters $\phiv$ together
(one step of $\thetav$ update and one step of $\phiv$ update within each mini-batch).
One natural extension is to make the numbers of the two type of updates adjustable.
We thus set the parameter $n_{\thetav}$ which controls the number of steps of $\thetav$ update within each mini-batch,
and parameter $n_{\phiv}$ which controls the number of steps of $\phiv$ update following each step of $\thetav$ update.
Larger $n_{\thetav}$ can potentially make the samples of $\thetav$ less correlated.
Larger $n_{\phiv}$ can potentially make the proposal distribution more accurate.
We set $n_{\thetav}=10$ and $n_{\phiv}=1$ in our implementation.


Finally, in Alg.~\ref{algo:dsmcmc-detail} we summarize a detailed version of our method.


\section{Derivations}
We provide the derivations of the Gibbs sampler for the \methodg{}.
The hidden variables are sampled layer-wisely and dimension-wisely.
We define $\zv^{(0)}=\xv$ and $\zv^{(L+1)}=\zerov$ for convenience.
Then the probability $p(z^{(l)}_{i}|\zv^{(l)}_{\neg i},\xv,\zv^{(\neg l)})$ can be written as $p(z^{(l)}_{i}|\zv^{(l)}_{\neg i},\zv^{(\neg l)})$.
We have the following Gibbs sampler:\\[-.5cm]
\begin{align*} 
   & p(z^{(l)}_{i}|\zv^{(l)}_{\neg i},\zv^{(\neg l)})  \\
 = & p(z^{(l)}_{i}|\zv^{(l)}_{\neg i},\zv^{(< l)}, \zv^{(> l)})  \\
\propto & p(\zv^{(< l)}|z^{(l)}_{i},\zv^{(l)}_{\neg i},\zv^{(> l)}) \cdot p(z^{(l)}_{i}|\zv^{(l)}_{\neg i},\zv^{(> l)})  \\
 = & p(\zv^{(< l)}|\zv^{(l)}) \cdot p(z^{(l)}_{i}|\zv^{(l+1)}) \\
\propto & p(\zv^{(l-1)}|\zv^{(l)}) \cdot p(z^{(l)}_{i}|\zv^{(l+1)}) \\
= &\!\! \prod_{i'=1}^{D^{(l-1)}}\!\!\!\!\!\exp\!\biggl[(\Wv^{(l-1)^\top}_{i,:}\!\!\!\zv^{(l)}\!+\!b^{(l)}_{i'})z^{(l-1)}_{i'} \!\!-\!\log(1\!+\!e^{(\Wv^{(l-1)^\top}_{i,:}\!\!\zv^{(l)}\!+\!b^{(l)}_{i'})})\biggr]\\
&\times\!\!\exp\!\biggl[(\Wv^{(l)^\top}_{i,:}\!\!\!\zv^{(l+1)}\!+\!b^{(l)}_{i})z^{(l)}_{i}\!\!-\!\!\log(1\!+\!e^{(\Wv^{(l-1)^\top}_{i,:}\!\!\!\zv^{(l+1)}+b^{(l-1)}_{i})})\biggr] \\
\propto & \exp\biggl[(\sum_{i'=1}^{D^{(l-1)}}W^{(l-1)}_{i'i}z^{(l-1)}_{i'}+(\Wv^{(l)^\top}_{i,:}\zv^{(l+1)}+b^{(l)}_{i}))z^{(l)}_{i}\\
&~~~~~~~~~~~~~-\sum_{i'=1}^{D^{(l-1)}}\log(1+e^{(\Wv^{(l-1)^\top}_{i',:}\zv^{(l)}+b^{(l)}_{i'})})\biggr].
\end{align*}\\[-.5cm]

A Gibbs sampler for \darn{} can be derived similarly.


\end{document}